\documentclass[journal]{IEEEtran}
\usepackage{amsmath,amsfonts}
\usepackage{algorithmic}
\usepackage{algorithmic}
\usepackage{algorithm}
\usepackage{array}
\usepackage[caption=false,font=footnotesize,labelfont=rm,textfont=rm]{subfig}% 20240102修改
\usepackage{textcomp}
\usepackage{stfloats}
\usepackage{url}
\usepackage{verbatim}
\usepackage{graphicx}
\usepackage{cite}
\usepackage{multirow}
\usepackage{subfig}
\usepackage{booktabs}
\usepackage{tabularx}
\usepackage{amsfonts}
\newcommand{\etal}{{\em et al\,. }}       % et al.
\newcommand{\eg}{{\em e.g., }}           % e.g.
\newcommand{\ie}{{\em i.e., }}           % i.e.
\newcommand{\etc}{{\em etc. }}         % etc.
\usepackage{makecell}

\usepackage{xcolor}

\begin{document}

\title{Unpaired Multi-view Clustering via Reliable View Guidance}

\author{Like~Xin, Wanqi~Yang*, Lei~Wang, Ming~Yang*
\thanks{This work is supported by the National Natural Science Foundation of China (Nos. 62276138, 62076135, 61876087), the Qing Lan Project of Jiangsu Province, China, and Postgraduate Research \&
Practice Innovation Program of Jiangsu Province. Ming Yang and Wanqi Yang are the co-corresponding authors.}
\thanks{Like Xin is with the School of Mathematical Sciences, Nanjing Normal University, Nanjing, 210046, China.  \protect (e-mail: xinlike94@gmail.com).}
\thanks{Wanqi Yang and Ming Yang are with the School of Computer and Electronic Information, Nanjing Normal University, Nanjing, 210046, China. \protect (e-mail: yangwq@njnu.edu.cn, myang@njnu.edu.cn).}% <-this % stops a space
\thanks{Lei Wang is with the School of Computing and Information Technology, University of Wollongong, Australia. \protect (e-mail: leiw@uow.edu.au).}
}

\markboth{IEEE TRANSACTIONS ON NEURAL NETWORKS AND LEARNING SYSTEMS}%
{Shell \MakeLowercase{\textit{et al.}}: Unpaired Multi-view Clustering via Reliable View Guidance}


\maketitle

\begin{abstract}
This paper focuses on unpaired multi-view clustering (UMC), a challenging problem where paired observed samples are unavailable across multiple views. The goal is to perform effective joint clustering using the unpaired observed samples in all views. In incomplete multi-view clustering, existing methods typically rely on sample pairing between views to capture their complementary. However, that is not applicable in the case of UMC. Hence, we aim to extract the consistent cluster structure across views. In UMC, two challenging issues arise: uncertain cluster structure due to lack of label and uncertain pairing relationship due to absence of paired samples. We assume that the view with a good cluster structure is the reliable view, which acts as a supervisor to guide the clustering of the other views. With the guidance of reliable views, a more certain cluster structure of these views is obtained while achieving alignment between reliable views and other views. Then we propose Reliable view Guidance with one reliable view (RG-UMC) and multiple reliable views (RGs-UMC) for UMC. Specifically, we design alignment modules with one reliable view and multiple reliable views, respectively, to adaptively guide the optimization process. Also, we utilize the compactness module to enhance the relationship of samples within the same cluster. Meanwhile, an orthogonal constraint is applied to latent representation to obtain discriminate features. Extensive experiments show that both RG-UMC and RGs-UMC outperform the best state-of-the-art method by an average of 24.14\% and 29.42\% in NMI, respectively.

\end{abstract}


\section{Introduction}

% \IEEEPARstart {I}{n} the real world, data often have diverse characteristics, collected from multiple sensors or obtained through various feature extractors, forming \textbf{multi-view data} \cite{xu2013survey, yang2015mrm}, \eg multimedia data with text, image, and audio, multi-modal neuroimaging data including Magnetic Resonance Imaging (MRI) and Computed Tomography (CT), and multilingual news story, \etc 
\IEEEPARstart {I}{n} {real-world scenarios, data frequently exhibit diverse characteristics, originating from various sensors or acquired through different feature extractors, resulting in multi-view data \cite{10149819, scl-UMC, xu2013survey, yang2015mrm}, \eg multimedia data including text, image, and audio, multi-modal neuroimaging data involving magnetic resonance imaging (MRI) and computed tomography (CT), and multilingual news stories, \etc \cite{scl-UMC}.}
% For example, it encompasses multimedia data comprising text, images, and audio, as well as multi-modal neuroimaging data such as Magnetic Resonance Imaging (MRI) and Computed Tomography (CT), and multilingual news stories, among others \cite{scl-UMC}. 
% However, due to various unpredictable or uncertain factors such as data noise, privacy protection, equipment malfunctions, or attribute degradation, several samples may be missing in different views, thereby resulting in \textbf{incomplete multi-view data} \cite{2015Multi}. 
{However, various factors like noise, privacy constraints, or equipment malfunctions \cite{10149819, scl-UMC} may lead to missing samples across different views, resulting in \textbf{incomplete multi-view data} \cite{2015Multi}.}
%As illustrated in Fig. \ref{unpaired} (b), solid shapes depict observed samples, while hollow shapes denote missing samples. It's important to note that some matched samples are still observed between views. 
Unfortunately, there exists an extreme yet realistic scenario where observed samples do not match between views \cite{10149819}, termed as \textbf{unpaired multi-view data} \cite{10149819,scl-UMC}. %As depicted in Fig. \ref{unpaired} (c), there are no common samples observed in both views. This type of multi-view data is referred to as \textbf{unpaired multi-view data}. In reality, \textbf{unpaired multi-view data} is common in various scenarios.
{For example, in a multi-camera surveillance system, individual cameras may operate intermittently due to factors like energy conservation or maintenance, leading to unpaired multi-view data. Similarly, web data from routers in different countries often lacks correspondence relationships \cite{2013Multi}. The frequent occurrence of sensor damage and replacements in multi-sensor detection systems also leads to unpaired multi-view data.}

{Clustering, a powerful learning method for revealing data cluster structure \cite{yang2018multi}, partitions samples into different clusters without supervision. In both complete multi-view data and incomplete multi-view data mentioned above, two downstream clustering tasks \textbf{multi-view clustering} and \textbf{incomplete multi-view clustering} \cite{scl-UMC,10149819}, arise.} These tasks have been extensively studied \cite{2018Doubly, 2013Multi, yang2018multi, wang2023efficient, li2021consensus}.

{Similarly, a downstream clustering task on unpaired multi-view data, referred to as \textbf{unpaired multi-view clustering} (UMC) \cite{10149819}, is more efficient when performing joint clustering across all views rather than conducting individual clustering within each view \cite{scl-UMC}.}
% Similarly, we aim to perform clustering on unpaired multi-view data, \ie \textbf{unpaired multi-view clustering} (UMC), since collecting the data from all views for joint clustering is more efficient than individual clustering in each view. 
{However, UMC is a more challenging task than multi-view clustering and incomplete multi-view clustering for the absence of paired samples \cite{scl-UMC}, which makes it a relatively understudied problem.}
Another issue, which may appear similar but distinct from UMC, arises when the paired relationships of inter-view samples are unknown \cite{yu2021novel, huang2020partially, yang2021MvCLN, yang2022robust, lin2022tensor}. In this case, different views establish paired relationships through recoupled self-representation matrices \cite{lin2022tensor}. However, in UMC, there are no paired samples between views. As a result, paired relationships cannot be constructed between views, making it impossible to address the UMC issue in our work.

% \re{Another issue, `uncoupled multi-view clustering', arises when the paired relationships of inter-view samples are unknown \cite{yu2021novel, huang2020partially, yang2021MvCLN, yang2022robust,lin2022tensor}. It appears similar but is different from UMC. In `uncoupled multi-view clustering', there still exist complete observed samples, and different views establish connections through recoupled self-representation matrices. However, in UMC, where complete observed samples do not exist, recoupled self-representation matrices cannot be constructed, making it impossible to address the UMC issue in our work.}

% {A similar issue arises \cite{yu2021novel, huang2020partially, yang2021MvCLN, yang2022robust} when the paired relationships of inter-view samples are unknown. These methods aim to mine the unknown paired relationships for multi-view clustering. % \xlk{data arrangements of different views are unknown.}
% The issue can be viewed as a form of incomplete multi-view clustering, given the existence of paired samples in multiple views. However, in UMC, there are no paired samples between observed samples from different views, resulting in a lack of pairing relationships. This is the most significant distinction between these two issues.}

% 每个视图有各自的聚类结构，观察发现视图之间的聚类结构具有一致性。因此，考虑用聚类结构建立视图之间的关系。

In multi-view clustering, different views exhibit consistent cluster structures. Even though UMC lacks paired samples, the cluster structures across different views remain consistent. {Therefore, we aim to investigate the consistency in cluster structures, as it is advantageous for both view matching and clustering.} 
{However, UMC encounters two significant challenges. Firstly, \textbf{uncertain cluster structure}: There is no label information on each view. Samples are assigned to clusters based on their distances from various cluster centroids. Consequently, clustering assignments can be easily altered without the guidance of supervised information. Secondly, \textbf{uncertain matching relationship}: Clusters across different views do not explicitly correspond to specific categories, making it challenging to establish matching relationships between them.} %To solve the two issues mentioned above, we propose a novel method, namely reliable KL divergence for unpaired multi-view clustering (RG-UMC). For the first issue, we appoint the reliable views in multi-view to guide these unreliable views to learn a better cluster structure. For the second issue, we incorporating KL divergence between the views with better cluster structure and other views with worse cluster structure, which efficient for us to assess alignment between different views and guide the clustering process accordingly. 

% 后期可用
% Fortunately, we found an effective way to solve these two issues. There are similar cluster structures among different views, while the clustering performance on each view is different. To relieve the uncertain cluster structure on each view, we adaptively leverage reliable views with perfect cluster structure to guide the cluster structure learning of other views. Besides, to solve the uncertain pairing relationship of clusters between views, we leverage KL divergence to achieve alignment between different views.
% We aim to enhance the determinism of the learned cluster structure and learn a consistent cluster structure across different views. 

{We strive to improve the reliability of the learned cluster structure and ensure its consistency across different views. }Meanwhile, we found the cluster performance on each view is different. Some views perform well in cluster structure, while others exhibit poor. Naturally, we want to explore an interesting question: \emph{whether the view with a good cluster structure assists in the learning of the view with a worse cluster structure?} 
% We found the clustering performance of individual views varies. 
% There are consistent cluster structures among different views, while the clustering performance on each view is different. %Surprisingly, we discovered that views with a better cluster structure can guide the learning of other views, thereby reducing the uncertainty in cluster structure formation and pairing relationships.
In identification system \cite{yang2015auxiliary}, fingerprints are considered a `strong' view as they provide more reliable information for recognizing the ID, while features like face descriptors and gaits are considered `weak' views for less accurate identification. Inspired by this, we define a \textbf{reliable view} as a supervisor in cluster structure compared to other views. Furthermore, we assume \textbf{the cluster structure of the reliable view acts as a supervisor to guide the clustering of the other views}, then cluster structure and pairing relationships are determined by reliable views. %Hence, the complementary information of cluster structure between views will be constructed. % Some related works leverage the strong view to guide the learning of the weak view. For semi-supervised learning, \cite{muslea2003active} and \cite{yang2015auxiliary} utilized strong views as soft supervision to assist weak views. For the unsupervised scenario, Lin \cite{lin2022tensor} fixed one view  as the reliable view to align others.
% For the works related to `reliable’, \eg `safe view clustering’\cite{tang2022deep} and `reliable view clustering’\cite{tao2018reliable}. They are used to improve the performance reliability, while the definition of `reliable’ in our work is the view {with good cluster structure}, thus the purpose of `reliable’ is different. 
Note that the effect of our reliable view is different from existing works of `reliable view clustering’, \eg \cite{tang2022deep} and \cite{tao2018reliable}. They are used to improve the performance reliability, while we aim to find a view with good cluster structure.

\begin{figure}[!t]
\centering
% \includegraphics[width=1.0 \columnwidth]{the_guidance_of_reliable_view.eps}
% \includegraphics[width=1.0 \columnwidth]{the_guidance_of_reliable_view_20230615.eps}
%\includegraphics[width=1.0 \columnwidth]{the_guidance_of_reliable_view_20230615.eps}
% \includegraphics[width=1.0 \columnwidth]{the_guidance_of_reliable_view1.eps}
% % \includegraphics[width=1.0 \columnwidth]{the_guidance_of_reliable_view_20230620.eps} 
% \includegraphics[width=1.0 \columnwidth]{the_guidance_of_reliable_view_20230620-2.eps} 
% \includegraphics[width=0.8 \columnwidth]{the_guidance_of_reliable_view_20230620-3.eps} % 3views
% \includegraphics[width=0.8 \columnwidth]{the_guidance_of_reliable_view_20230620-4.eps} % OK
\includegraphics[width=1.0 \columnwidth]{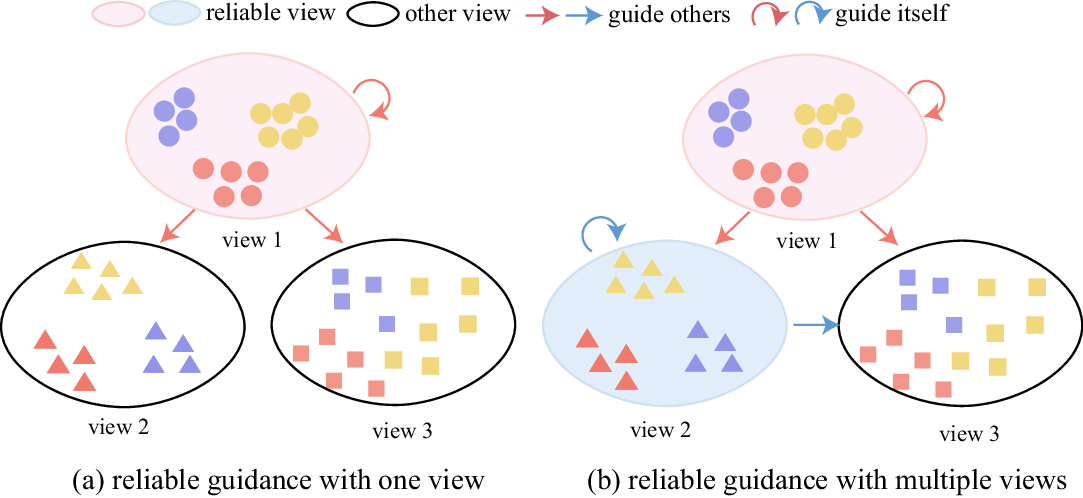}
% 3views
% \vspace{-0.3cm}
% \caption{The guidance strategy of reliable view with single view (in red) and multiple views (in blue)}\label{reliable}
% \caption{The strategies of reliable view guidance with one view (in red) and multiple views (in blue). Assuming the reliable ranking of three views is as follows: view 1 $>$ view 2 $>$ view 3. In one reliable view guidance (in red), view 1 is designated as the reliable view, guiding both its cluster structure learning and that of other views (view 1 and view 2). In the multiple reliable views guidance (in blue), views with stronger cluster structures serve as directors to guide views with weaker cluster structures. \eg view 1 and view 2 both guide the learning of view 3. The strategies fully leverage all reliable views to effectively align cluster structures across different views.}\label{reliable}
\caption{The strategies of reliable view guidance with one view and multiple views. Assuming the reliable ranking of three views is as follows: view 1 $>$ view 2 $>$ view 3. In (a), view 1 is designated as the reliable view, guiding both its cluster structure learning and that of other views (view 1 and view 2). In (b), views with stronger cluster structures serve as directors to guide views with weaker cluster structures. \eg view 1 and view 2 both guide the learning of view 3. The strategies fully leverage all reliable views to effectively align cluster structures across different views.}\label{reliable}
% \vspace{-0.1cm}
\end{figure}

For the utilization of a reliable view, a similar work was proposed \cite{lin2022tensor}, which aligns other views with a fixed single view. However, the method has the following limitations: i) The appointed reliable view remains constant, overlooking the possibility of other reliable views emerging during the iterations; ii) The single reliable view neglects the guiding potential of other sub-optimal views.
% For the utilization of reliable view, a similar work was proposed \cite{lin2022tensor}, which fixed one view as the reliable view to align others. However, the approach of relying on a fixed single view for guidance has limitations: \xlk{i) For the fixed reliable view, the appointed reliable view remains constant. Therefore, it overlooks the possibility of other reliable view emerging during the iterations.} ii) For the one reliable view, it neglects the guiding potential of other sub-optimal views. 
Considering these limitations, we design two strategies of reliable view guidance in the unsupervised setting: one reliable view (for the fixed view) and multiple reliable views (for one reliable view) in each iteration, as illustrated in Fig \ref{reliable}. %\re{Both of these strategies adaptively designate reliable views during the iterations.} 
% \re{The reliable views could be changed during the iterations. Therefore, both of these strategies adaptively designate reliable views.} 
During the iterations, the designation of reliable views can adaptively change based on the reliability of these views. 
% Besides, the strategies fully utilize all reliable views and effectively align the cluster structures of different views under the guidance of reliable views, thereby an improvement in clustering performance.
{Furthermore, these strategies fully utilize all reliable views and effectively align cluster structures across views under the guidance of reliable views, thereby an improvement in clustering performance.}

Therefore, to solve the aforementioned issues (uncertain cluster structure and uncertain matching relationships), we designate the reliable view as a supervisor to guide the learning of cluster structures in other views. Meanwhile, the other views are matched with the reliable view, resulting in increased certainty of both cluster structure and pairing relationships.
% Therefore, to solve the aforementioned issues (uncertain cluster structure and uncertain matching relationship), we designate the reliable view, which acts as a supervisor to guide the cluster structure learning of others, meanwhile the other views match to the reliable view, resulting in increased certainty of cluster structure and pairing relationships.
%\xlk{With the aid of reliable view guidance, the cluster structures of multiple views are learned by the reliable view, meanwhile the other views match to the reliable view, resulting in increased certainty of cluster structure and pairing relationships.} % The certainty of cluster structure increase, as the reliable view with good cluster structure. The certainty of pairing relationship increases for multiple views are matching to reliable views. Afterwards, we design two approaches with one and multiple reliable views guidance, respectively. Both of them choose reliable views adaptively during optimization.
Specifically, we design alignment modules for reliable view guidance with one and multiple views, respectively. Both modules adaptively select reliable views during optimization.
Besides, we employ a compactness module to strengthen the relationships among samples within the same cluster. An orthogonal constraint is also applied to the latent representation to extract discriminative features. These modules collectively enhance the clustering performance. Therefore, two novel methods called Reliable view Guided Unpaired Multi-view Clustering with one view (RG-UMC) and multiple views (RGs-UMC) are proposed. %In sum, our main contributions can be summarized as:
% \begin{itemize}
% \item We investigate a challenging but insufficiently studied problem, unpaired multi-view clustering, where no observed paired samples exist between any two views. To establish relationships between views and achieve alignment, we explore guidance from reliable views based on consistent cluster structures.
{The primary contributions can be outlined as follows:}

\begin{itemize}
\item {We address the challenging and relatively unexplored problem of UMC, which arises when no paired samples are available between any two views. To establish relationships between views and achieve alignment, we explore guidance from reliable views based on consistent cluster structures.}%across views. 
% To establish relationships between views, we explore their consistent and complementary information on cluster structure by reliable view.
% We study the problem of unpaired multi-view clustering, where no paired samples exist between any two views. %information by reliable view guidance.
%To establish relationships between different views, we utilize views with reliable cluster structures to guide other views.
\item By utilizing the guidance of reliable views, we propose two novel methods to learn the effective cluster structure for UMC with one reliable view guidance (RG-UMC) and multiple reliable views guidance (RGs-UMC), both of which adaptively specify the reliable views. %By utilizing the guidance of reliable views, we propose two novel methods to learn the effective cluster structure for UMC with one reliable view guidance (RG-UMC) and multiple reliable views guidance (RGs-UMC). For RG-UMC, one view is specified as the reliable view to guide the learning process dynamically. While for RGs-UMC, multiple reliable views are used to guide adaptively.
% \item Compared with the sixteen state-of-the-art methods, extensive experiments on multi-view benchmark datasets show excellent performance. 
\item {Experiments on five datasets validate superior performance compared to sixteen methods.} Specifically, RG-UMC outperforms the best comparison method with average improvements of 24.14\% (NMI), 37.49\% (ACC), and 33.05\% (F-score). Similarly, RGs-UMC achieves higher performance with average improvements of 29.42\% (NMI), 41.36\% (ACC), and 35.47\% (F-score). %Besides, we conduct ablation studies, convergence analysis, and t-SNE visualization to validate the efficacy of the proposed models. %For comprehensive exploration of the inter-view relationships, RGs-UMC outperforms RG-UMC in performance and accelerates the formation of cluster structure.
% \xlk{Besides, we proved that RGs-UMC accelerates the formation of cluster structure than RG-UMC in experiment for comprehensive exploration of the inter-view relationships.} 
% Due to the comprehensive exploration of the inter-view relationships, RGs-UMC outperforms RG-UMC in terms of performance.

\end{itemize}

% \begin{figure}
% % \centering
% \includegraphics[width=1.0 \columnwidth]{the_guidance_of_reliable_view.eps}
% \vspace{-0.5cm}
% \caption{The guidance strategy of reliable view with single view and multiple views}\label{reliable}
% \vspace{-0.3cm}
% \end{figure}
%appoint single view as reliable view.

% Therefore, the strategies make full use of all reliable views.

\section{Related Work}\label{relatedwork}

\subsection{Incomplete Multi-view Clustering} 
In real-world scenarios, missing samples in one or more views can lead to incomplete observations, posing a challenge for clustering tasks. For incomplete multi-view clustering, some researchers have explored various methods, primarily including spectral approaches and subspace learning methods \cite{10149819, scl-UMC}. %For spectral approaches, Rai \etal \cite{rai2010multiview} proposed a kernel CCA-based multi-view clustering method that completes the kernel matrix using Laplacian and kernel matrices from complete views. However, this method relied on the assumption that at least one view is complete. 
In spectral approaches, Rai \etal \cite{rai2010multiview} presented a kernel CCA-based method for multi-view clustering, which completes the kernel matrix using Laplacian and kernel matrices from complete views. However, this method assumes the availability of at least one complete view. Wen \etal \cite{Jie2019Unified} leverage a locality-preserving term to infer missing views for natural alignment across all views. %Additionally, it employs reverse graph regularization to adaptively learn a consensus graph, ensuring consistency in the local structure across multiple views. 
% Besides, Wen \etal \cite{2020Jie} proposed a framework that integrates graph construction and consensus representation learning. It employs low-rank representation, spectral constraint, and co-regularization to optimize similarity graphs for each view and obtain a shared cluster representation across views. 
Besides, Wen \etal \cite{wen2021unified} proposed a tensor spectral clustering method to recover missing views and utilize both the hidden information of these missing views and the intra-view information of data, which is often overlooked in current approaches. {For the incomplete multi-view data, recovering the original high-dimensional data is time-consuming and noise-sensitive, and separating the cluster indicator learning into a separate step may lead to sub-optimal results. Therefore, Zhang \etal \cite{zhang2023robust} proposed a method that integrates spectral embedding completion and discrete cluster indicator learning into a unified framework to address these issues.}

For subspace learning approaches, Yin \etal \cite{2017Unified} devised a method for incomplete multi-view subspace learning, which enhances performance by simultaneously addressing feature selection and similarity preservation within and across views. % To fully leverage observed samples, Yang \etal \cite{2018Incomplete} utilized sparse low-rank representation learning for each view to impute missing samples within a view. Subsequently, they learned a common subspace to establish connections between views, facilitating downstream clustering tasks.
To fully leverage observed samples, Yang \etal \cite{2018Incomplete} utilizes a combined sparse and low-rank matrix to model the correlation among samples within each view to impute missing samples. Moreover, it enforces similar subspace representations to explore the relationships between samples across different views, facilitating downstream clustering tasks. Thanks to the coefficient matrix, the self-representation method reflects the precise relationships among samples. Therefore, Liu \etal \cite{liu2021self} introduced a self-representation subspace clustering algorithm tailored for incomplete multi-view scenarios, which elegantly integrates completion of missing samples and self-representation learning into a cyclical process. %which elegantly integrates missing data imputation and self-representation learning in a cyclical process. 
With a unified sparse subspace learning framework, Li \etal \cite{li2023anchor} proposed a method to learn inter-view anchor-to-anchor and intra-view anchor-to-incomplete affinities. 
{Then these components are integrated to generate a unified clustering outcome through a consensus sparse anchor graph. However, obtaining matched samples across all views, as required by these methods, can be challenging in real-world scenarios.} % These are then combined into a consensus sparse anchor graph to produce a unified clustering result. However, these methods need available samples matched across all views, which can be challenging to obtain in real-world scenarios.

% However, these above methods require some observed samples matched among all views, which may be relatively demanding in real scenarios.
% \subsection{Contrastive Learning for Multi-view Clustering} % 对比学习通常需要增广样本，而多视图学习中天然存在
% The contrastive learning methods have shown strong power in unsupervised representation learning. The main idea of contrastive learning is maximizing the similarities of positive pairs while minimizing the negative pairs in a feature space. 

% To date, some multi-view clustering works with contrastive learning have been fully explored. Trosten \etal \cite{trosten2021reconsidering} added a contrastive learning component, which is a selective alignment procedure that preserves the model’s ability to prioritize views. For the limitation of the same feature space explored in existing works, Xu \etal \cite{xu2022multi} proposed Multi-level Feature Learning for Contrastive Multi-view Clustering (MFLVC), which learns different levels of features from the raw features, including low-level features, high-level features, and semantic labels/features in a fusion-free manner. For the large-scale online learning scenarios, Li \etal \cite{li2021contrastive} proposed an online clustering method called Contrastive Clustering (CC), which explores the contrastive learning on instance-level and cluster-level. Though some multi-view learning works with contrastive learning have been explored \cite{li2021contrastive, trosten2021reconsidering, yang2022robust, xu2022multi}, they commonly require some complete sample to share. 

\subsection{Unpaired Multi-view Clustering} 
% Unpaired multi-view clustering (UMC) represents an extreme scenario of incomplete multi-view clustering, with each sample observed in only one view. Therefore, UMC would be more difficult than incomplete multi-view clustering. 
Unpaired multi-view clustering (UMC) is an extreme scenario of incomplete multi-view clustering (IMC), where samples are observed in just one view, making it more challenging than IMC \cite{10149819,scl-UMC}.
Several researchers have utilized weak supervision information to establish correlations between views. For instance, both Qian \etal \cite{2013Multi} and Houthuys \etal \cite{2017unpaired} incorporated must-link and cannot-link constraints to construct correlations across views in their respective works. Nonetheless, these methods may not be effective in the absence of paired samples or supervised information. 
{For UMC with no supervision, Yang \etal \cite{10149819} proposed IUMC aimed at learning consistent subspace representations for clustering. Based on the baseline, two unpaired multi-view clustering methods were developed. Specifically, IUMC-CA aligns covariance matrices to enhance subspace consistency before clustering, whereas IUMC-CY directly clusters using assignments, offering efficiency for large datasets.}
Besides, Xin \etal \cite{scl-UMC} proposed a method for UMC (scl-UMC), which leverages inner-view and inter-view selective contrastive learning modules to enhance the certainty of cluster structures. However, scl-UMC emphasizes the consistency between views, overlooking the potential complementary that exists among them.
% However, the alignment between views has mainly focused on local cluster structure alignment, neglecting the importance of global alignment. Additionally, in unpaired multi-view clustering, the emphasis has been solely on exploring the consistency between views, overlooking the potential complementarity that exists among them. To address these limitations, we propose novel methods, RG-UMC and RGs-UMC, which dynamically leverage the cluster structures of reliable views to guide the learning process and capture complementary information between views. By considering both local and global alignment, our models aim to enhance the performance of unpaired multi-view clustering tasks.
% 其他相关工作

% Although only a few methods existed in UMC, we also investigate a similar issue where the mapping relationship of inter-view samples is unknown, due to privacy protection, \emph{etc}. 
% Besides, We concern a related issue wherein the mapping relationship between inter-view samples remains unknown. 
{Additionally, we are also concerned about the related issue of unknown mapping relationships between inter-view samples \cite{huang2020partially, yu2021novel, yang2021MvCLN, yang2022robust}.}
%Yu \etal \cite{yu2021novel} utilized graph structure consistency in each view to learn the cross-view sample mapping matrix by non-negative matrix factorization. 
{For example, through non-negative matrix factorization, Yu \etal \cite{yu2021novel} utilized graph structure consistency to construct the mapping matrix between cross-view samples.}
% Huang \etal \cite{huang2020partially} proposed using a differentiable surrogate for the non-differentiable Hungarian algorithm to establish category-level correspondence between views when the mapping relationship between inter-view samples is partially unknown. 
% Huang \etal \cite{huang2020partially} proposed a method that utilizes a differentiable surrogate for the non-differentiable Hungarian algorithm to establish category-level correspondence between views in scenarios where the mapping relationship between inter-view samples is only partially known. 
Yang \etal \cite{yang2021MvCLN} tackled the issue using robust contrastive learning. Moreover, in their subsequent work a more robust multi-view clustering SURE \cite{yang2022robust} was introduced, which directly re-aligns and recovers samples from other views. 
% These methods strive to find sample-matching relationships between views, either in the original space or subspace. However, such relationships are often absent in UMC.
These methods attempt to establish sample-matching relationships across views. However, in UMC, such relationships are often lacking. Therefore, our aim is to explore consistency in clustering.

\subsection{Multi-view learning via view guidance}
% Strong modality guide the learning of weak modality.
% the cluster structure of strong modality acts like a supervisor, to guide the weak  modality clustering. 
% \cite{tao2018reliable}
% \cite{muslea2003active}
% \cite{yang2015auxiliary}
% \cite{yang2016learning}
% \cite{pmlr-v29-Wang13b}
% \cite{yangyang2020reliable}
% Besides, the labels of weak modal instances also be considered.
%\cite{muslea2006active} extend the multi-view learning framework by also exploiting weak views, which are adequate only for learning a concept that is more general/specific than the target concept.

% For supervised and semi-supervised scenarios, Muslea \etal \cite{ muslea2006active} proposed co-Testing, which extend the multi-view learning framework by exploiting strong and weak views in multi-view learning. 

{In supervised and semi-supervised scenarios, Muslea \etal \cite{ muslea2006active} introduced co-testing, an extension of the multi-view learning framework that leverages both strong and weak views in multi-view learning. }%They utilized the disagreement between strong and weak classifiers to provide additional labeling information. 
Previous co-training methods assumed that each view can independently provide accurate predictions \cite{yangyang2020reliable}, while the assumption is unrealistic due to potential feature corruption or noise. 
%Then Wang \etal \cite{wang2013co} further showed that co-training could improve learning performance even with insufficient views, as long as the views have significant diversity. %For multi-view learning via view guidance on supervised and semi-supervised scenarios, Muslea \etal \cite{muslea2003active, muslea2006active} introduced co-Testing, an approach that considered strong and weak views. They assumed that the concentrated examples whose labels from strong classifiers are different and inconsistent with the prediction of the weak classifier provide more information for labeling. Besides, previous theoretical analyses of co-training methods assumed that each view is independently sufficient for accurate predictions\cite{yangyang2020reliable}. However, this assumption is unrealistic due to feature corruption or noise, then Wang \etal \cite{wang2013co} proved that if the two views have large diversity, co-training was able to improve the learning performance by exploiting unlabeled data even with insufficient views. %Besides, the importance of different modalities is various in real tasks, while previous works assumed that each modality contains sufficient information for the target and can be treated with equal importance. Therefore, for semi-supervised multi-view learning, using a weak modality with limited labeled data to build a classifier can be challenging in real-world scenarios. Then Yang \etal \cite{yang2015auxiliary} proposed the Auxiliary information Regularized Machine (ARM) for classification.
For the importance of different modalities varies in real tasks, Yang \etal \cite{yang2015auxiliary} proposed the auxiliary information regularized machine (ARM) model, which leveraged auxiliary information from a strong modality to guide feature extraction on the weak modality.
% Besides, the importance of different modalities varies. In semi-supervised multi-view learning, using a weak modality with limited labeled data for classification is challenging. Therefore, Yang \etal. proposed the Auxiliary information Regularized Machine (ARM) model, which leverages auxiliary information from a strong modality to guide feature extraction on the weak modality. % Besides, the importance of different modalities is various in real tasks. For semi-supervised multi-view learning, using a weak modality with limited labeled data to build a classifier can be challenging in real-world scenarios. Then Yang \etal \cite{yang2015auxiliary} proposed the Auxiliary information Regularized Machine (ARM) for classification. The model performed feature extraction on weak modal features using auxiliary information from the strong modality as supervision. %which aims to improve the multi-modal learning performance while extracting the most discriminative weak modal feature subspace at the same time. %It also improves the classification performance of the strong modality by leveraging the weak modal features as a regularization component. 
Furthermore, considering the cost of strong modal feature extraction, Yang \etal \cite{yang2016learning} proposed the active querying strong modalities (ACQUEST) training strategy, which actively queried the strong modal feature values of `selected’ instances instead of relying on their corresponding ground truths. Additionally, in semi-supervised scenarios, %Yang \etal \cite{yang2019semi} considered the instance-level auto-encoder for a single modality and modified bag-level optimal transport to strengthen the consistency among modalities.
{Yang \etal \cite{yang2019semi} investigated the utilization of instance-level auto-encoder for individual modalities and a modification of bag-level optimal transport to improve consistency across modalities. Besides, in the specific application of Corporate Relative Valuation, Yang \etal \cite{yang2021corporate} solved by a heterogeneous multi-modal graph neural network.}
% Besides, in semi-supervised scenarios, Yang \etal \cite{yang2019semi} employed extrinsic unlabeled data by incorporating instance-level auto-encoder and ensuring bag-level consistency across various unlabeled modal predictions using the modified OT metric. % to learn the modal consistency on bag-level prediction, Yang \etal \cite{} considers the bag-level consistency among different unlabeled modal predictions with the modified OT theory.
Although these methods have achieved good performance, they need some supervision information to assist. 

% \cite{cui2021self, mi2024fast,wang2021fast, chen2023incomplete}
% self-guidance; anchor guidance; complete view guidance; reliable view guidance.
% In unsupervised learning, as the deep subspace clustering network underestimates the significance of view-fusion, Cui \etal \cite{cui2021self} proposed a self-supervised model for simultaneous subspace clustering, consensus construction, and self-guided learning. To learn the consensus to be more clustering-friendly, they designed spectral supervisors by the self-guidance of pseudo labels. 
{In unsupervised learning, Cui \etal \cite{cui2021self} proposed a self-supervised model that performs simultaneous subspace clustering, consensus construction, and self-guided learning. To enhance clustering friendliness, they introduced spectral supervisors through self-guided pseudo-labeling.} % Considering the heterogeneous gap between modalities, Yang \etal \cite{yang2022exploiting} transformed both the raw image and corresponding generated sentence into the shared semantic space and measured the generated sentence from prediction consistency and relation consistency. 
{Given the heterogeneity between modalities, Yang \etal \cite{yang2022exploiting} bridged the gap by aligning both raw images and corresponding generated sentences into a shared semantic space. They assessed the generated sentences based on prediction consistency and relational consistency.} %As multi-view subspace clustering methods suffer from high time costs and are challenging to use in real-life large-scale data, Mi \etal \cite{mi2024fast} and Wang \etal \cite{wang2021fast} employed anchors guidance to select crucial landmarks, effectively reducing time consumption. 
{Multi-view subspace clustering methods present challenges in real-world large-scale data applications due to the high time costs and complexity. Therefore, Mi \etal \cite{mi2024fast} and Wang \etal \cite{wang2021fast} employed anchors guidance to select crucial landmarks, effectively reducing time consumption. } Considering the incomplete multi-view data is not missing in any view, Chen \etal \cite{chen2023incomplete} designed a knowledge distillation framework and proposed an incomplete multi-view clustering with complete view guidance. {Lin \etal \cite{lin2022tensor} introduced T-UMC for addressing incomplete coupling among different views in multi-view data by identifying the most reliable view for alignment. }
%To solve the issue of the multi-view data not completely coupled between different views, Lin \etal \cite{lin2022tensor} proposed a tensor approach for uncoupled multiview clustering (T-UMC) to identify the most reliable view for coupling. %Specifically, the coupling relationships between views were established using a view-specific coupling matrix, and the t-SVD-based tensor low-rank approximation was leveraged to capture higher-order correlations among all views.
However, T-UMC used a fixed single view as the reliable view, which limits the guidance from potentially better views during optimization and overlooks the guidance from other suboptimal views. Additionally, it did not consider the clustering task with missing samples, particularly in the extreme case of UMC with no paired samples.

\section{Methodology}\label{proposedmethod}
% \subsection{Notations}
\subsection{Notations and Architecture}
% For a clear and precise description, we first introduce notations of the main symbols used in this paper, which is also listed in TABLE \ref{table1} to facilitate checking their meanings. %Generally, a bold upper case letter denotes a matrix, a bold lower case letter denotes a vector, an upper case decorated letter represents a set, and a normal lower case letter corresponds to a scalar. The superscript indicates the view index, while the subscript indicates the sample index. 
% \textbf{Notations.} We assume that there are $V$ different views. Unpaired multi-view data $\{\boldsymbol{X}^v\}_{v=1}^V$ are derived from the $V$ views, where $v$ varies within the range $\{1,2,..,V\}$. For the $v$-th view, there are $n^v$ observed samples that form the observed sample matrix $\boldsymbol{X}^v\in \mathbb{R}^{ n^v \times d^v}$, \ie $\boldsymbol{X}^v=[\boldsymbol{x}_1^v;\boldsymbol{x}_2^v;\ldots ;\boldsymbol{x}_{n^v}^v]$, and each sample has $d^v$-dimensional features. Note that, in the unpaired case, all the $n^v$ observed samples cannot be matched with those of all the other views. Thus, all views together have $N$ different samples totally, \ie $N=\sum_{v=1}^V n^v$. 

\textbf{Notations.} Unpaired multi-view data $\{\boldsymbol{X}^v\}_{v=1}^V$ stem from $V$ views, where $v$ within $\{1,2,..,V\}$. In the $v$-th view, the observed sample matrix $\boldsymbol{X}^v\in \mathbb{R}^{ n^v \times d^v}$ consists of $n^v$ samples $\boldsymbol{X}^v=[\boldsymbol{x}_1^v;\boldsymbol{x}_2^v;\ldots ;\boldsymbol{x}_{n^v}^v]$, with each sample possessing $d^v$-dimensional features in the $v$-th view. The total number of observed samples in unpaired multi-view data is denoted as $N$, where $N=\sum_{v=1}^V n^v$ due to the absence of paired samples between views. {At the beginning of training, the observed samples $\boldsymbol{X}^v$ in the $v$-th view are inputted into their corresponding autoencoder, which consists of an encoder $F^v$ and a decoder $G^v$ respectively. Subsequently, the latent subspace representation $\boldsymbol{Z}^v$ obtained by $F^v(\boldsymbol{X}^v)$.}
% At the beginning of training, the observed samples on each view $\boldsymbol{X}^v$ are fed into its specific autoencoder, where the encoder and the decoder in the $v$-th view are denoted as $F^v$ and $G^v$, respectively. Then we obtain the latent subspace representation $\boldsymbol{Z}^v$ by $F^v(\boldsymbol{X}^v)$. % We learn the latent subspace representations $\{\boldsymbol{Z}^v\}_{v=1}^V$ for clustering by using the widely-used autoencoder in all views, where the encoder and the decoder in the $v$-th view are denoted as $F^v$ and $G^v$, respectively. 
% \xlk{For the alignment module with reliable views, we set $r$ as the index for one reliable view, and apply matrix $\boldsymbol{W}$ denote the weight of multiple reliable views, respectively.}
The clustering performance of a view can be regarded as the reliability of that view in a clustering task. Here, we use the silhouette coefficient to compute the reliability of a view. Specifically, $sil(\boldsymbol{z}_i^v)$ represents the silhouette coefficient of sample $\boldsymbol{z}_i^v$, and $sils^v$ denotes the silhouette coefficient of the $v$-th view.
% {For the silhouette coefficient, $sil(\boldsymbol{z}_i^v)$ represents the silhouette coefficient of data point $\boldsymbol{z}_i^v$, and $sils^v$ denotes the silhouette coefficient of the $v$-th view. } 
% and $sils^v$ denotes the average silhouette coefficient of all samples in the $v$-th view.  %For the silhouette coefficient, $sil(\boldsymbol{z}_i^v)$ is the silhouette coefficient of data point $\boldsymbol{z}_i^v$, $sils^v$ is the silhouette coefficient of $v$-th view. 
In alignment modules with reliable views, we set $r$ as the index for one reliable view, and apply matrix $\boldsymbol{W}$ to denote the weight of multiple reliable views, respectively. %\xlk{To improve clarity, we include the main symbols used in this paper in the supplementary material.}
For clarity, the primary symbols are detailed in the supplementary material.

\begin{figure*}[!t]
\centering
\includegraphics[width=1.8\columnwidth]{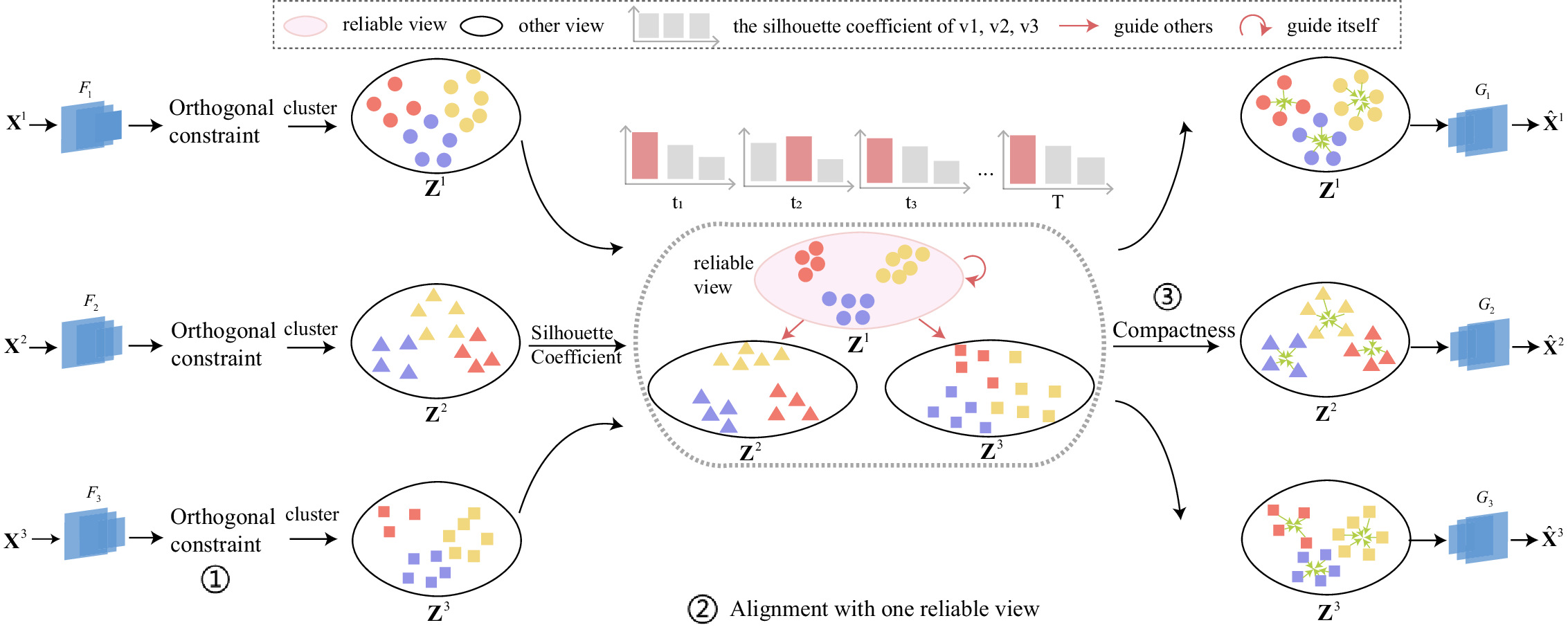}% 3views; single reliable view guidance
% \vspace{-0.3cm}
\caption{{The RG-UMC framework illustrates the process for a batch dataset with three views.} The model is designed to achieve a clear and consistent cluster structure across multiple views. It consists of three essential components: an orthogonal constrain, an alignment module with one reliable view, and a compactness module. In the alignment module, the most reliable view dynamically changes with the arrival of different batches of data.}
% \caption{\xlk{The framework demonstrates the procedure for a batch dataset. }The RG-UMC and RGs-UMC models are designed to achieve a clear and consistent cluster structure across multiple views. They consist of three essential components: an orthogonal module, an alignment module with one reliable view, and a compactness module. The main difference between the two models is the number of reliable views (single-view guidance or multiple-view guidance) used in the reliable KL divergence module.} 
% The framework of RG-UMC and RGs-UMC. The models are designed to learn the clear and consistent cluster structure among views. Specifically, they are consist of an orthogonal module, an alignment module  with reliable views, and a compactness module. The difference between RG-UMC and RGs-UMC models is reliable KL divergence module with single-view guidance or multiple-view guidance.
\label{KL-framework}
% \vspace{-0.1cm}
\end{figure*}

\textbf{The architecture of RG-UMC.}
As depicted in Fig. \ref{KL-framework}, we employ autoencoder networks in each view to extract view-specific latent representations, which are utilized for clustering subsequently \cite{scl-UMC}. To enhance the performance, we introduce a learning process for the latent representations involving orthogonality constrain, an alignment module, and a compactness module. %we introduce a learning process for the latent representations involving modules such as orthogonality, \xlk{alignment with one reliable view} and compactness. 

% RG-UMC and RGs-UMC 同时介绍
%In our approach, RG-UMC and RGs-UMC, we employ autoencoder networks in each view to extract view-specific latent representations, which are utilized for clustering subsequently. To enhance the performance of multi-view clustering, we introduce a learning process for the latent representations involving modules such as orthogonality, \xlk{alignment with reliable view} and compactness. The architecture of our models is depicted in Fig. \ref{KL-framework-one-multiple}. \xlk{Specifically, the alignment module has two styles with one reliable view and multiple reliable views.}
%Specifically, the alignment module has two guiding strategies with single-view guidance and multiple-view guidance.
% reliable KL divergence,
% We simply take three views for example. Specifically, in Fig. \ref{KL-framework}, given a set of three-view samples ($\boldsymbol{X}^1, \boldsymbol{X}^2, \boldsymbol{X}^3$), 
% We simply take three views for example. Specifically, in Fig. \ref{KL-framework}, given a batch of three-view samples ($\boldsymbol{X}^1, \boldsymbol{X}^2, \boldsymbol{X}^3$), they are fed through the three encoders ($F^1, F^2, F^3$) to learn view-specific subspace representations ($\boldsymbol{Z}^1, \boldsymbol{Z}^2, \boldsymbol{Z}^3$). Then, in the subspace, ($\boldsymbol{Z}^1, \boldsymbol{Z}^2, \boldsymbol{Z}^3$) are processed by three modules to learn the consistent subspace representations between views. 

{We illustrate the process with three views. As depicted in Fig. \ref{KL-framework}, a batch of samples from three views ($\boldsymbol{X}^1, \boldsymbol{X}^2, \boldsymbol{X}^3$) fed into their corresponding encoders ($F^1, F^2, F^3$) to generate three subspace representations ($\boldsymbol{Z}^1, \boldsymbol{Z}^2, \boldsymbol{Z}^3$). Subsequently, these subspace representations are processed by three modules in the subspace to learn consistent representations across views.}
%Then, in the subspace, ($\boldsymbol{Z}^1, \boldsymbol{Z}^2, \boldsymbol{Z}^3$) are processed by an orthogonal constrain, \xlk{an alignment module with one reliable view,} and a compactness module to learn the consistent subspace representations between views. 
% After convergence, the subspace representation ($\boldsymbol{Z}^1, \boldsymbol{Z}^2, \boldsymbol{Z}^3$) are used to recover raw feature representation ($\hat{\boldsymbol{X}^1}, \hat{\boldsymbol{X}^2}, \hat{\boldsymbol{X}^3}$) by the three decoders ($G^1, G^2, G^3$), respectively. Finally, the learned subspace representations from views are integrated into a consistent representation matrix $\boldsymbol{Z}=[\boldsymbol{Z}^1; \boldsymbol{Z}^2; \boldsymbol{Z}^3]$ to obtain the final clustering assignments. 
{Once convergence, the subspace representations ($\boldsymbol{Z}^1, \boldsymbol{Z}^2, \boldsymbol{Z}^3$) are employed to reconstruct raw feature representations ($\hat{\boldsymbol{X}^1}, \hat{\boldsymbol{X}^2}, \hat{\boldsymbol{X}^3}$) via their respective decoders ($G^1, G^2, G^3$). Subsequently, these subspace representations from different views are combined into a unified matrix $\boldsymbol{Z}=[\boldsymbol{Z}^1; \boldsymbol{Z}^2; \boldsymbol{Z}^3]$, facilitating the final clustering assignments \cite{scl-UMC}.} Specifically, before feeding data into the alignment module with one reliable view, we need to calculate the index $r$ for the reliable view using the silhouette coefficient. As shown in Fig. \ref{KL-framework}, only the view with the highest silhouette coefficient guides the learning of other views in the alignment module.
% RG-UMC and RGs-UMC 同时介绍
% Specifically, before feeding data into the alignment module with reliable views, we need to calculate the index $r$ or weight matrix $W$ for the reliable view using the silhouette coefficient. 
% Furthermore, in the alignment module with reliable views, two strategies are designed. In the alignment with one reliable view, only the view with the highest silhouette coefficient guides the learning of other views. Conversely, in the alignment with multiple reliable views, all views except the one with the lowest silhouette coefficient contribute to guiding the cluster structure learning of other views. 

%Furthermore, in the reliable KL divergence module, two guiding strategies are designed. In the single-view guidance strategy, only the view with the highest silhouette coefficient, \eg $\boldsymbol{Z}^1$, guides the learning of other views. Conversely, in the multiple-view guidance strategy, all views except the one with the lowest silhouette coefficient, \eg$\boldsymbol{Z}^3$, contribute to guiding the cluster structure learning of other views. 

\begin{figure*}[!t]
\centering
\includegraphics[width=1.8\columnwidth]{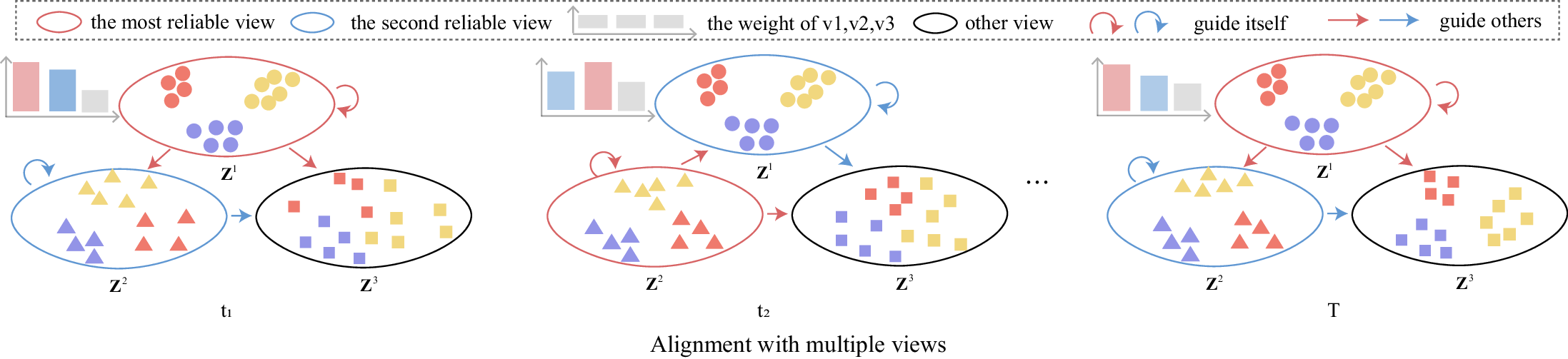}
% \vspace{-0.1cm}
\caption{The alignment module of RGs-UMC involves three views. The reliable view dynamically changes based on the silhouette coefficient in each batch during round $T$ optimization. The strategy fully leverages all reliable views to effectively align cluster structures across different views.}% Additionally, each view's numerical values, representing silhouette coefficients after normalization, sum to 1 across multiple views.
% \caption{\xlk{The alignment module of RGs-UMC with three views. It's worth noting that the reliable view undergoes significant changes during round $T$ optimization. In each view, the numerical values represent silhouette coefficients after normalization, and the sum of these values across multiple views equals 1. During the optimization, the reliable view is specified dynamically according to the silhouette coefficient in current batch.}}%All views, except the one with the lowest silhouette coefficient, play a role in guiding the cluster structure learning of other views.
\label{KL-framework-multiple}
% \vspace{-0.5cm}
\end{figure*}

\textbf{The architecture of RGs-UMC.} %Considering the views with suboptimal silhouette coefficients may also assist in forming a consistent cluster structure and alignment, RGs-UMC is designed. 
The only difference between RG-UMC and RGs-UMC is the number of reliable views in the alignment module. Consequently, we depict the alignment module of RGs-UMC in Fig. \ref{KL-framework-multiple}. Similar to RG-UMC, we calculate the weight matrix $W$ for the reliable view using the silhouette coefficient before feeding the data into the alignment module. Throughout the $T$ iterative optimizations, the reliable view and its corresponding weight adaptively change based on the silhouette coefficient. As depicted in Fig. \ref{KL-framework-multiple}, most views can act as supervisors guiding the learning of other views.
% As shown in Fig. \ref{KL-framework-multiple}, all views, except for the one with the lowest silhouette coefficient, contribute to guiding the learning of cluster structures in other views. 
% In the single-view guided strategy (Fig. \ref{KL-framework} (b)), only the view with the most reliable cluster structure ($\boldsymbol{Z}^1$) guides the learning of other views. In contrast, in the multiple-views guided strategy (Fig. \ref{KL-framework} (c)), all views, except the most unreliable one ($\boldsymbol{Z}^3$), guide the cluster structure learning of other views. % Specifically, the reliable KL divergence module has two strategies to guide. For the single-view guidance strategy in Fig. \ref{KL-framework} (b), only view with the most reliable cluster structure ($\boldsymbol{Z}^1$) guide the learning of others. Conversely, for the multiple-view guidance strategy in Fig. \ref{KL-framework} (c), except the most unreliable view ($\boldsymbol{Z}^3$), all views are guide the cluster structure learning of others. % Specifically, the reliable KL divergence module has two strategies to guide. For the single-view guidance strategy in Fig. \ref{KL-framework} (b), only view with the most reliable cluster structure guide the learning of others. Conversely, for the multiple-view guidance strategy in Fig. \ref{KL-framework} (c), except the most unreliable views, all views are guide the cluster structure learning of others with reliable lower than itself. 

% \subsection{The Autoencoder and Silhouette Coefficient}
\textbf{Multi-view Autoencoders with regularizer.}
% Autoencoder \cite{xu2022multi,scl-UMC} is a widely used unsupervised model, which can project the raw features into a latent feature space to learn view-specific information. Specifically, an autoencoder consists of an encoder and a decoder, where the encoder projects the raw representations to the low-dimensional space, and the decoder maps the low-dimensional representations to the raw space. Then, for the feature representation $\boldsymbol{X}^v$ in the \emph{v}-th view, we pass it through an autoencoder to learn the latent representation $\boldsymbol{Z}^v$ by minimizing the reconstruction loss. 
% \xlk{The autoencoder \cite{xu2022multi,scl-UMC} is a common unsupervised model that aims to project raw features into a latent feature space to capture view-specific information. Specifically, it comprises an encoder and a decoder: the encoder transforms raw representations into a low-dimensional space, while the decoder maps low-dimensional representations back to the raw space. 
The autoencoder is a common unsupervised model in multi-view learning, aiming to capture view-specific information by projecting raw features into a low-dimension space \cite{xu2022multi,scl-UMC}. Commonly, an autoencoder comprises an encoder and a decoder. %Thus, for the feature representation $\boldsymbol{X}^v$ in the $v$-th view, we utilize an autoencoder to learn its latent representation $\boldsymbol{Z}^v$ by minimizing the reconstruction loss \cite{scl-UMC}.
Therefore, we utilize an autoencoder to learn the latent representation $\boldsymbol{Z}^v$ of the features $\boldsymbol{X}^v$ in the $v$-th view, minimizing the reconstruction loss \cite{scl-UMC}.
%Besides, an orthogonal constraint on the latent representation $\boldsymbol{Z}^v$ is used to avoid pushing the integral space arbitrarily \cite{chen2022adaptively} and restrict the representation to be more discriminative \cite{chen2022efficient}. 
Furthermore, applying an orthogonal constraint to the latent representation $\boldsymbol{Z}^v$ prevents it from expanding arbitrarily within the integral space \cite{chen2022adaptively} and enforces a more discriminative representation \cite{chen2022efficient}. {Subsequently, to learn the latent representation $\boldsymbol{Z}^v$, we utilize an autoencoder with a regularizer \cite{scl-UMC} to process $\boldsymbol{X}^v$ by minimize the following loss:}
% Then, for the feature representation $\boldsymbol{X}^v$ in the \emph{v}-th view, we pass it through an autoencoder with regularizer \cite{scl-UMC} to learn the latent representation by minimizing the following loss: %minimizing the reconstruction loss and regularizer loss as follows:
\begin{equation}\label{eq1}
\setlength\abovedisplayskip{1.5pt}
\setlength\belowdisplayskip{1.5pt}
% \small
l_{AE} = \sum_{v=1}^V ||\boldsymbol{X}^v - G^v\big(F^v(\boldsymbol{X}^v)\big)||_F^2 + \lambda_1 ||\boldsymbol{Z}^v {\boldsymbol{Z}^v}^{T} - \boldsymbol{I}_{d}||_F^2,%+ \lambda_1 ||\boldsymbol{Z}^v {\boldsymbol{Z}^v}^{T} - \boldsymbol{I}_{d}||_F^2,
\\
\end{equation}
where $F^v$ and $G^v$ are the encoder and decoder of the \emph{v}-th autoencoder. Specifically, we define $\boldsymbol{Z}^v$ as the latent representation of $\boldsymbol{X}^v$, \ie $\boldsymbol{Z}^v = F^v(\boldsymbol{X}^v)$. %Besides, an orthogonal constraint on $\boldsymbol{Z}^v$ is used to avoid pushing the integral space arbitrarily \cite{chen2022adaptively} and restrict the latent representation to be more discriminative \cite{chen2022efficient}. 
{$\lambda_1$ is the hyperparameter to balance reconstruction loss and regularizer loss.}
% The formula can be written as:
% \begin{equation}\label{eq2}
% \setlength\abovedisplayskip{1.5pt}
% \setlength\belowdisplayskip{1.5pt}
% \small
% l_{Orth} = \sum_{v=1}^V ||\boldsymbol{Z}^v {\boldsymbol{Z}^v}^{T} - \boldsymbol{I}_{d}||_F^2,
% \\
% \end{equation}
% \subsection{Proposed Method}
\subsection{RG-UMC: UMC via One Reliable View Guidance}
\textbf{The silhouette coefficient in multiple views.} 
%To measure the cluster structure performance of single view, we leverage silhouette coefficient to measure
% The silhouette coefficient\cite{rousseeuw1987silhouettes, lin2022tensor} measures the distance between objects with their own cluster and other clusters. With cluster method, \eg $K$-means, we get the cluster assignment on each view. Then, based on these clustering assignmnets, we have the following two values for a data point $z_i$
As in \cite{lin2022tensor}, we leverage the silhouette coefficient to select reliable views. The silhouette coefficient \cite{rousseeuw1987silhouettes, lin2022tensor} is utilized to evaluate the clustering performance of each view. This metric assesses the distance between samples belonging to the same cluster and those in different clusters \cite{lin2022tensor}. %, which measures the distance between samples within their own cluster and other clusters \cite{lin2022tensor}. 
Therefore, we firstly obtain the cluster assignment on each view with a cluster method, \eg $K$-means. Then, based on these cluster assignments, we calculate the two values $a(\boldsymbol{z}_i^v)$ and $b(\boldsymbol{z}_i^v)$ for the sample $\boldsymbol{z}_i^v$ in latent space. Certainly, %Definitely, 
$a(\boldsymbol{z}_i^v)$ represents the average distance between $\boldsymbol{z}_i^v$ and other samples within the same cluster, while $b(\boldsymbol{z}_i^v)$ is the minimum average distance between $\boldsymbol{z}_i^v$ and samples from different clusters. Specifically, the formulation is as follows:
\begin{align}
\setlength\abovedisplayskip{1.5pt}
\setlength\belowdisplayskip{1.5pt}
&a(\boldsymbol{z}_i^v) = \frac{1}{|\Omega_k^v|-1} \sum_{\boldsymbol{z}_j^v \in \Omega_k^v, j \neq i} s_{ij}^v,\\
&b(\boldsymbol{z}_i^v) = \min\limits_{k \neq k'} \frac{1}{|\Omega_{k'}^v|} \sum _{\boldsymbol{z}_j^v \in \Omega_{k'}^v} s_{ij}^v,
\end{align}
% \begin{equation}\label{eq3}
% \setlength\abovedisplayskip{1.5pt}
% \setlength\belowdisplayskip{1.5pt}
% a(\boldsymbol{z}_i^v) = \frac{1}{|\Omega_k^v|-1} \sum_{\boldsymbol{z}_j^v \in \Omega_k^v, j \neq i} s_{ij}^v,
% \end{equation}
% \begin{equation}\label{eq4}
% \setlength\abovedisplayskip{1.5pt}
% \setlength\belowdisplayskip{1.5pt}
% \small
% % b(\boldsymbol{z}_i^v) = \min\limits_{k \neq k'} \frac{1}{|\Omega_{k'}^v|} \sum _{j=1}^{ |\Omega_{k'}^v|} s_{ij}^v,
% b(\boldsymbol{z}_i^v) = \min\limits_{k \neq k'} \frac{1}{|\Omega_{k'}^v|} \sum _{\boldsymbol{z}_j^v \in \Omega_{k'}^v} s_{ij}^v,
% \end{equation}
% where {$\Omega_k^v$ is the set of samples in the $k$-th cluster of the $v$-th view, }$s_{ij}^v = ||\boldsymbol{z}_i^v - \boldsymbol{z}_j^v||_2^2$ is the Euclidean distance between sample $\boldsymbol{z}_i^v$ and $\boldsymbol{z}_j^v$, $\boldsymbol{z}_i^v$ is assigned to $\Omega_k^v$, $|\Omega_k^v|$ and $|\Omega_{k'}^v|$ are the number of samples in clusters $k$ and $k'$ in the $v$-th view, respectively. Then the silhouette coefficient for sample $\boldsymbol{z}_i^v$ is defined as follows:
{where $\Omega_k^v$ represents the sample set in the $k$-th cluster of the $v$-th view, while $s_{ij}^v = ||\boldsymbol{z}_i^v - \boldsymbol{z}_j^v||_2^2$ calculates the Euclidean distance between samples $\boldsymbol{z}_i^v$ and $\boldsymbol{z}_j^v$, where $\boldsymbol{z}_i^v$ belongs to $\Omega_k^v$. The number of samples in clusters $k$ and $k'$ in the $v$-th view are denoted as $|\Omega_k^v|$ and $|\Omega_{k'}^v|$, respectively. Then, the silhouette coefficient for sample $\boldsymbol{z}_i^v$ is defined as:}
\begin{equation}\label{eq5}
% \setlength\abovedisplayskip{1.5pt}
% \setlength\belowdisplayskip{1.5pt}
% \small
\begin{split}
sil(\boldsymbol{z}_i^v) = \left \{
\begin{array}{ll}
    \frac{b(\boldsymbol{z}_i^v)-a(\boldsymbol{z}_i^v)}{{\rm max}(a(\boldsymbol{z}_i^v), b(\boldsymbol{z}_i^v))},                    & \textit{{\rm if}}  \quad |\Omega_k^v|> 1,\\
    0,                    & \textit{{\rm if}}  \quad |\Omega_k^v|= 1.\\
\end{array}
\right.
\end{split}
\end{equation}
As shown in Eq. (\ref{eq5}), the average silhouette coefficient $sil(\boldsymbol{z}_i^v)$ varies from -1 to 1. %A higher value of $sil(\boldsymbol{z}_i^v)$ indicates that samples within the same cluster are more similar to each other and more distinct from samples in different clusters. % \xlk{A higher value of $sil(\boldsymbol{z}_i^v)$ suggests that samples exhibit greater similarity to each other within the same cluster and are more dissimilar from samples in different clusters.}
{A higher value of $sil(\boldsymbol{z}_i^v)$ suggests that samples exhibit greater similarity to each other within the same cluster while dissimilar from samples in different clusters.}

% A higher $sil(\boldsymbol{z}_i^v)$ value indicates that samples within the same cluster are closer to each other while samples from different clusters are farther apart.

Multi-view clustering aims to leverage observed samples in each view for effective joint clustering. %However, it is difficult to distinguish which view is more suitable for the clustering task \cite{lin2022tensor}. 
However, choosing the most suitable view for the clustering task presents a challenging issue \cite{lin2022tensor}. {The clustering performance of a view can be considered as its reliability in a clustering task.} %The clustering performance of a view can be regarded as the reliability of that view in a clustering task.
% {Therefore, the reliability of $v$-th view computed by silhouette coefficient is defined as follows:}
{Therefore, the reliability of the $v$-th view calculated by the silhouette coefficient as follows:}
% Multi-view clustering aims to leverage observed samples in each view for effective joint clustering. However, the cluster performance on each view is different, a view with good cluster performance would more reliable to help other views improve. Therefore, it is important to distinguish which view is more suitable for the clustering task \cite{lin2022tensor}. As for the clustering task, a view with good cluster structure can be viewed as the reliable view to guide other views. We regard the clustering performance of a view as the reliability of that view. Then,  the reliability of $v$-th view computed by silhouette coefficient is defined as follows:
\begin{equation}\label{eq7}
\setlength\abovedisplayskip{1.5pt}
\setlength\belowdisplayskip{1.5pt}
% \small
sils^v = \frac{1}{n^v} \sum_{i=1}^{n^v} sil(\boldsymbol{z}_i^v), 
\end{equation}
%we simplify the silhouette coefficient of $v$-th view as $sils^v$,
exactly, the value range of $sils^v$ is the same as that of $sil(\boldsymbol{z}_i^v)$, which is [-1, 1]. A higher silhouette coefficient value for $v$-th view $sils^v$ indicates that the view exhibits better cluster structure.
% The silhouette coefficient can assess the quality of the data clustering. 

The silhouette coefficient serves as a measure to evaluate the clustering quality of the data. For selecting the reliable view, we suppose the view with the highest silhouette coefficient is the reliable view \cite{lin2022tensor}. Then, we get the index of the most reliable view $r$ as follows:
%with the help of $silhouette coefficient$, we choose a view with the highest silhouette coefficient as the most reliable view $r$:
\begin{equation}\label{eq6}
% \setlength\abovedisplayskip{1.5pt}
% \setlength\belowdisplayskip{1.5pt}
% \small
r = \max \limits_v sils^v, 
\end{equation}
% where
% \begin{equation}\label{eq7}
% \setlength\abovedisplayskip{1.5pt}
% \setlength\belowdisplayskip{1.5pt}
% \small
% sils^v = \frac{1}{n^v} \sum_{i=1}^{n^v} sil(\boldsymbol{z}_i^v), 
% \end{equation}
where $sils^v$ denotes the silhouette coefficient of $v$-th view in Eq. (\ref{eq7}), which can be easily computed according to \cite{rousseeuw1987silhouettes}.

% \textbf{Orthogonal constraint.}

% \textbf{RG-UMC.}
\textbf{Alignment with one reliable view.}
In UMC, establishing relationships between views is challenging for the absence of paired samples. Fortunately, the consistent cluster structure across views allows us to construct relationships between views. As the reliability of different views varies in multi-view learning, we leverage the view with a reliable cluster structure to guide other views with a relatively unreliable cluster structure, thereby improving the performance of all views. 

{For align the cluster structure, there are many methods to achieve it, \eg contrastive learning \cite{trosten2021reconsidering}, mutual information maximization \cite{lin2021completer}, KL divergence \cite{buchner2022intuition}, \etc We choose one of them KL divergence for distribution alignment.} %Conveniently, we leverage KL divergence for alignment.
Subsequently, we design {the alignment module with reliable views by KL divergence} to achieve alignment between reliable views and the remaining views, thereby mitigating the uncertain cluster structure and uncertain matching relationship in UMC. 
%Subsequently, we design the reliable KL divergence module to achieve guidance and alignment between reliable views and the remaining views, to alleviate the uncertain cluster structure and uncertain matching relationship in UMC. %Then, we design an alignment module  with reliable views to align reliable view adaptively. 

Specifically, the alignment module with one reliable view assigns the reliable view firstly, then aligns other views with reliable view by KL divergence. % For {assigning the reliable view}, we design in two ways: \textbf{one reliable view} and \textbf{multiple reliable views}. % Specifically, the module \textbf{assigns the reliable view} firstly, then \textbf{align other views with reliable view} by KL divergence.
Therefore, we define the alignment module with one reliable view by KL divergence as follows:
% \begin{equation}\label{eq10}
% \setlength\abovedisplayskip{1.5pt}
% \setlength\belowdisplayskip{1.5pt}
% \small
% l_{KL} = \frac{1}{V}\sum_{v=1}^V D_{KL}\Big({\rm logsoftmax}(\boldsymbol{Z}^v )|| {\rm softmax}(\boldsymbol{Z}^{r})\Big), %, v \neq r 实验验证加入置信视图效果会提升
% \end{equation}
% where $\boldsymbol{Z}^r$ is the reliable view with the highest silhouette coefficient, $D_{KL}$ is the KL divergence \cite{buchner2022intuition}. %\cite{KL-divergence}. 
% \xlk{For the $\rm softmax()$ and $\rm logsoftmax()$, the information will be introduce follows. }
\begin{equation}\label{eq10}
\setlength\abovedisplayskip{1.5pt}
\setlength\belowdisplayskip{1.5pt}
% \small
% l_{KL} = \frac{1}{V}\sum_{v=1}^V D_{KL} (P^v || Q^r)
% l_{KL} = \frac{1}{V}\sum_{v=1}^V \sum_{i=1}^{n^v} P(z_i^v) {\rm log}(\frac{P(z_i^v)}{Q(z_i^r)}),
l_{KL} = \frac{1}{V}\sum_{v=1}^V P^v {\rm log}(\frac{P^v}{Q^r}), 
\end{equation}
% \re{Specifically, to adapt the `director' (reliable views) and `learner' (other views) for the KL divergence \cite{buchner2022intuition}, the samples from the `director' and `learner' need to be fed into the softmax layer and logsoftmax layer \cite{abdollahi2023nodecoder}, respectively. }
% \re{where $D_{KL}$ represents the KL divergence \cite{buchner2022intuition}.} 
where $r$ is the index of the reliable view. $P^v$ and $Q^r$ represent the probability distributions corresponding to the $v$-th view and the $r$-th reliable view, respectively. %These distributions can be calculated using a softmax layer \cite{abdollahi2023nodecoder, drainakis2020federated, niu2022spice}.
These distributions are approximated by the softmax layer applied to $\boldsymbol{Z}^v$ and $\boldsymbol{Z}^r$ \cite{abdollahi2023nodecoder, drainakis2020federated, niu2022spice}. $Q^r$ serves as the `director' (reliable view) to guide the cluster structure learning of $P^v$. Therefore, we achieve alignment between views through KL divergence, resulting in a consistent cluster structure across views. % Therefore, through KL divergence, we obtain a consistent cluster structure between views.

\textbf{Compactness.}
Compactness is a common evaluation used to assess the quality of clusters based on the similarity of samples within each cluster \cite{2014compactness}. %Thus, a good clustering will create clusters with samples that are similar or closest to one another. Besides, the lower value of the CP, the closer the intra cluster distance is. 
A good cluster structure exhibits lower compactness within each cluster. Therefore, to enhance the relationship of sample with same clustering, we define the compactness module in multi-view as:
% \begin{equation}\label{eq4}
% l_\mathcal{C} = \sum_{v=1}^V \overline{cp_i}^v,
% \end{equation}
% \begin{equation}\label{eq5}
% \overline{cp}^v = \frac{1}{K} \sum_{k=1}^K \frac{1}{\overline{cp}_k},
% \end{equation}
\begin{equation}\label{eq11}
\setlength\abovedisplayskip{1.5pt}
\setlength\belowdisplayskip{1.5pt}
% \small
l_\mathcal{C} = \frac{1}{VK}\sum_{v=1}^V \sum_{k=1}^K \frac{1}{\overline{\mathcal{C}}_k^v},
\end{equation}
where
\begin{equation}\label{eq12}
\setlength\abovedisplayskip{1.5pt}
\setlength\belowdisplayskip{1.5pt}
% \small
\overline{\mathcal{C}}_k^v = \frac{1}{|{\Omega}_k^v|}\sum_{\boldsymbol{z}_i \in {\Omega}_k^v} ||\boldsymbol{z}_i^v - \boldsymbol{c}_k^v||, 
\end{equation}
$\boldsymbol{z}_i^v$ and $\boldsymbol{c}_k^v$ are the $i$-th sample and the $k$-th cluster centriod in $v$-th view, $|\Omega_k^v|$ is the sample number in the $k$-th cluster of the $v$-th view.% $|\Omega_k^v|$ is the sample number of $k$-th cluster in $v$-th view.

% In sum, to learn the complementary information and the consistent cluster structure between views, 
% In sum, to solve the issues of uncertain cluster structure and uncertain matching relationship, the model \textbf{RG-UMC}, aligned by one reliable view in the UMC, is written by:
{In sum, addressing the challenges of uncertain cluster structure and uncertain matching relationships, the model \textbf{RG-UMC}, aligned by one reliable view in the UMC, is written by:}
\begin{equation}\label{eq13}
% \setlength\abovedisplayskip{1.5pt}
% \setlength\belowdisplayskip{1.5pt}
% \small
% L_{RG} = l_{AE}+  \lambda_1 l_{Orth} + \lambda_2 l_{KL}+ \lambda_3 l_\mathcal{C},\\
L_{RG} = l_{AE}+ \lambda_2 l_{KL}+ \lambda_3 l_\mathcal{C},\\
\end{equation}
where $l_{AE}$, $l_{KL}$, and $l_\mathcal{C}$ are three terms of autoencoder network with orthogonal constraint, alignment module with one reliable view and compactness module, respectively. Besides, $\lambda_2$ and $\lambda_3$ are the hyperparameters to balance these modules. 

\subsection{RGs-UMC: UMC via Multiple Reliable Views Guidance}
% \textbf{RGs-UMC} % the reliable view may change dramatically during iteration.
For the RG-UMC, the reliable view may change during iteration. Furthermore, we believe that the views with suboptimal silhouette coefficients may also have a positive effect on forming a consistent cluster structure and alignment. Therefore, we define the views with silhouette coefficient higher than current view as the reliable views and design the alignment module with multiple reliable views.
% A view with a higher silhouette coefficient helps with the formation of a cluster structure compared to a view with a lower silhouette coefficient. 
% \xlk{From the experiments, we observe that the reliable view changes dramatically in RG-UMC during iteration. Furthermore, we believe that the views with suboptimal silhouette coefficients may also have a positive effect on forming a consistent cluster structure and alignment. Therefore, we further design the alignment module with multiple reliable views. }
% \textbf{Alignment with multiple reliable views.} A view with a higher silhouette coefficient helps with the formation of a cluster structure compared to a view with a lower silhouette coefficient. Therefore, in alignment module with multiple reliable views, we define the views with silhouette coefficient higher than current view as the reliable views. 

\begin{figure}
\centering
% \vspace{-0.6cm}
\includegraphics[width=0.95 \columnwidth]{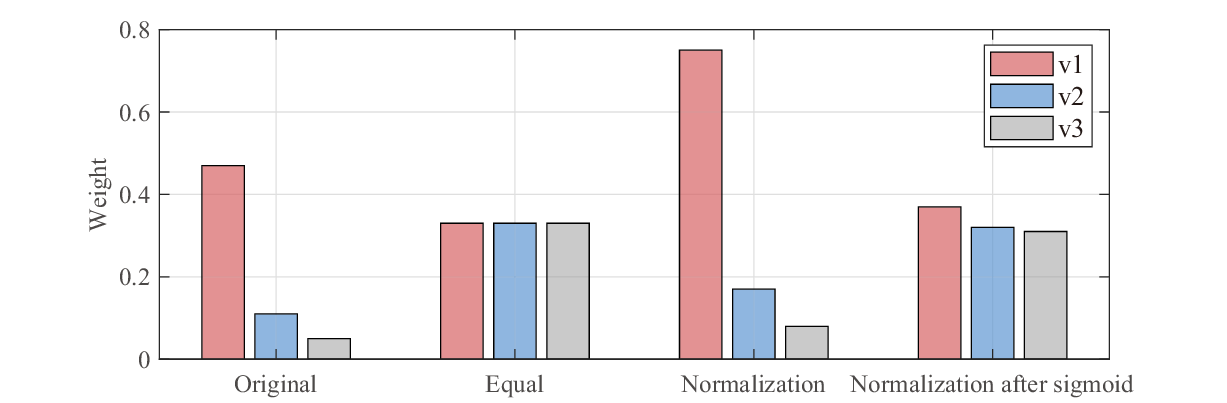}
\vspace{-0.2cm}
\caption{Taking three views as an example, the original weights are processed using three strategies: Equal, Normalization, and Normalization after sigmoid.}\label{sigmoid_function}
\vspace{-0.4cm}
\end{figure}
Additionally, as shown in Fig. \ref{sigmoid_function}, we observe that i) treating all views equally would reduce the effectiveness of views with higher silhouette coefficients (compare the columns `Original' and `Equal'). ii) Directly normalizing the silhouette coefficients across views would weaken the guidance from views with lower silhouette coefficients (compare the columns `Normalization' and `Normalization after sigmoid'). 
% Additionally, it is important to note that i) treating all views equally would reduce the effectiveness of views with higher silhouette coefficients. ii) Directly normalizing the silhouette coefficients across views would weaken the guidance from views with lower silhouette coefficients. %Besides, we must point that i) treating each view indiscriminately would decrease the effective of these views with higher silhouette coefficients. ii) Directly normalizing the silhouette coefficients across multiple views would diminish the guiding effect of views with lower silhouette coefficients. 
To address these issues, we employ sigmoid function to reprocess the silhouette coefficients on each view. % before normalization. 
Then the weight matrix of multiple reliable views formulate as follows:
% \begin{equation}
% \begin{split}
% w_{ij}= \left \{
% \begin{array}{ll}
%    \sigma(sil_j)/(\sum_{j=1}^V \sigma(sil_j)),                    & \textit{if}\quad sil_{j} \geq sil_{i},\\
%     0,                    & \textit{otherwise.}\\
% \end{array}
% \right.
% \end{split}\label{eq16}
% \end{equation}
\begin{equation}
% \setlength\abovedisplayskip{1.5pt}
% \setlength\belowdisplayskip{1.5pt}
% \small
\begin{split}
w_{vr}= \left \{
\begin{array}{ll}
   \sigma(sils^r)/\sum_{v=1}^{|\Omega_R|} \sigma(sils^v),                    & \textit{\rm {if}}\quad sil_{r} \geq sil_{v},\\%& \textit{\rm {if}}\quad  \rm{sils^{r}} \geq  sils^{v},\\
    0,                    & \textit {\rm{otherwise.}}\\
\end{array}
\right.
\end{split}\label{eq16}
\end{equation}
where $sils^r$ is the silhouette coefficient of the $r$-th view defined in Eq. (\ref{eq7}). $r$ is the view index with a higher silhouette coefficient than the $v$-th view. $\Omega_R$ and $|\Omega_R|$ are the set and number of these views indexes with higher silhouette coefficient. %The equation $\sigma(u)=\frac{1}{1+\exp(-u)}$ denotes it demonstrates rising trends gradually decreased. 
Besides, the sigmoid function $\sigma(u)$ rapidly increases initially and then gradually slows down. This property implies that even a minor silhouette coefficient can result in a relatively high weight, effectively smoothing the weights assigned to different views \cite{yin2003flexible}. Therefore, based on the Eq. (\ref{eq16}), we define the alignment module with multiple reliable views by KL divergence as follows:
\begin{equation}\label{eq10-2}
% \setlength\abovedisplayskip{1.5pt}
% \setlength\belowdisplayskip{1.5pt}
% \small
% l_{KLs} = \sum_{v=1}^V \sum_{r=1}^{|\Omega_R|} \frac{w_{vr}}{V^2} D_{KL}( P^v || Q^r),
l_{KLs} = \sum_{v=1}^V \sum_{r=1}^{|\Omega_R|} \frac{w_{vr}}{V^2} P^v {\rm log}(\frac{P^v}{Q^r})), 
\end{equation}
{where $|\Omega_R|$ is the number of indexes corresponding to views with a higher silhouette coefficient compared to $\boldsymbol{Z}^v$. Similar to Eq. (\ref{eq10}), $P^v$ and $Q^r$ are the approximate probability distribution corresponding to the $v$-th view and the $r$-th view.} %$P^v$ and $Q^r$ are calculated by ${\rm logsoftmax}(\boldsymbol{Z}^v)$ and ${\rm softmax}(\boldsymbol{Z}^r)$, respectively. %$r$ is the index of a reliable view with a higher silhouette coefficient compared to $\boldsymbol{Z}^v$. 
Moreover, instead of aligning with one reliable view, we leverage multiple views to align and further propose the \textbf{RGs-UMC} model as follows:
\begin{equation}\label{eq13-1}
% \setlength\abovedisplayskip{1.5pt}
% \setlength\belowdisplayskip{1.5pt}
% \small
% L_{RGs} = l_{AE}+  \lambda_1 l_{Orth} + \lambda_2 l_{KLs}+ \lambda_3 l_\mathcal{C},\\
L_{RGs} = l_{AE}+ \lambda_2 l_{KLs}+ \lambda_3 l_\mathcal{C},\\
\end{equation}
where $l_{KLs}$ denotes the alignment module with multiple reliable views. The hyperparameters $\lambda_2$ and $\lambda_3$ are utilized to balance these modules. The key difference between RG-UMC and RGs-UMC lies in the number of reliable views integrated into the alignment module.

\textbf{Implementation Details.}
% For the autoencoder we used, the detail of the network architecture is presented in TABLE \ref{table2_networkarchitecture}. 
% For the autoencoder we used, same to \cite{lin2021completer,scl-UMC}, we simply adopt a dense (\ie fully-connected) network where each layer is followed by a batch normalization layer and a ReLU layer. %Here, autoencoder adopts three same modules, where each module consists of a fully-connected network, a batch normalization layer, and a ReLU layer in order. 
% Furthermore, it must be pointed out that the softmax activation function is used in the last layer of the encoders.
% The detail of the network architecture is presented in supplementary. 
% For the autoencoder we used, same to \cite{lin2021completer,scl-UMC}, we simply adopt a dense (\ie fully-connected) network where each layer is followed by a batch normalization layer and a ReLU layer \cite{lin2021completer, scl-UMC}. %In the training stage, we adopt the algorithm of mini-batch gradient descent to train the RG-UMC model and RGs-UMC model. 
% \re{For the autoencoder used in this paper, it is same to \cite{lin2021completer,scl-UMC}.} \re{In the training stage, the mini-batch gradient descent is emploied to train both the RG-UMC and RGs-UMC models.} 
{The structure of the autoencoder used in this work is similar to that in \cite{lin2021completer,scl-UMC}.} Additionally, during training, we employ mini-batch gradient descent to train both the RG-UMC and RGs-UMC models. The unpaired data is used to train the whole network for 50 epochs. {Once the network converges, we use the saved autoencoder parameters to generate the latent representations $\{{\boldsymbol{Z}^v}\}_{v=1}^V$ for all views. Then, same to \cite{scl-UMC}, we conduct $K$-means clustering on the concatenated representations $\boldsymbol{Z}=[\boldsymbol{Z}^1; \boldsymbol{Z}^2; \ldots; \boldsymbol{Z}^V]$ to obtain the clustering assignments.}

% Once the network converged, we leverage the parameters saved by the autoencoders and feed the whole dataset into network to obtain the latent representation $\{\boldsymbol{Z}^v\}_{v=1}^V$ for all views. After that, we conduct $K$-means on the concatenation $\boldsymbol{Z}=[\boldsymbol{Z}^1; \boldsymbol{Z}^2; \ldots; \boldsymbol{Z}^V]$ to achieve the final clustering assignment.

\begin{figure}
\centering
\vspace{-0.6cm}
\includegraphics[width=0.9 \columnwidth]{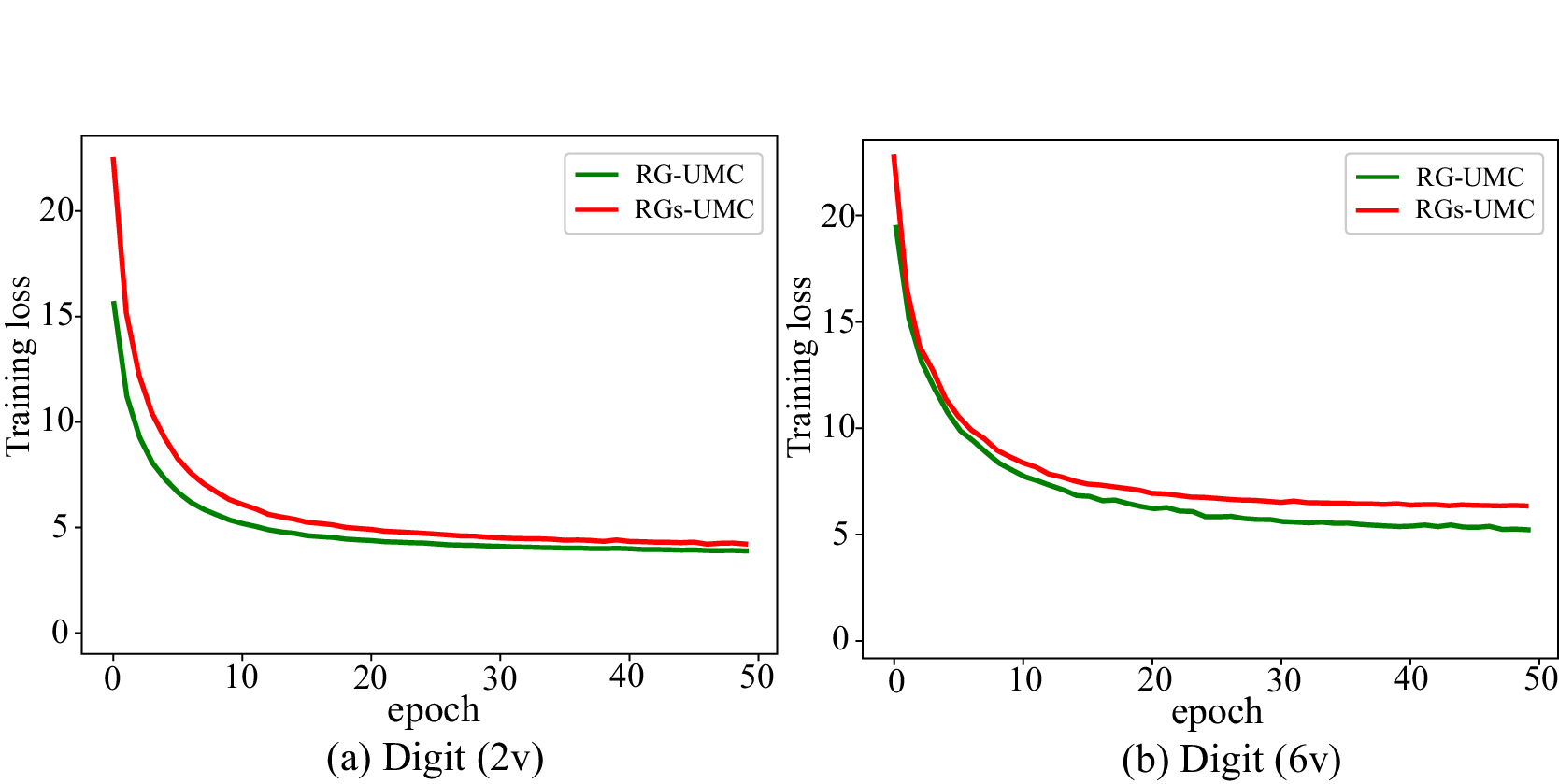} % 无reliable loss
\vspace{-0.2cm}
\caption{Training loss in RG-UMC and RGs-UMC.}\label{loss_Joint}
\vspace{-0.3cm}
\end{figure}

\textbf{Convergence analysis.} The complete algorithms for RG-UMC and RGs-UMC are outlined in Algorithm \ref{alg:algorithm1}. The distinctions between them can be observed in rows 6-7 and rows 9-10. %\xlk{The full algorithms of RG-UMC and RGs-UMC are provided in Algorithm \ref{alg:algorithm1} and Algorithm \ref{alg:algorithm2}, respectively.} 
Besides, we demonstrate the convergence of RG-UMC and RGs-UMC. Fig. \ref{loss_Joint} (a) and (b) depict the training loss of the two models on the \emph{Digit} dataset with two views and six views, respectively. %As observed, the loss of RG-UMC and RGs-UMC converges rapidly and steadily after 50 iterations. 
It can be observed that the loss of both RG-UMC and RGs-UMC converges rapidly and steadily after 50 iterations.

\begin{algorithm}[H]
\small
\caption{Proposed RG-UMC and RGs-UMC}\label{alg:algorithm1}
\textbf{Input:} unpaired multi-view data $\{ X^v\}_{v=1}^V$, the cluster number $K$, $epoch$, $batchsize$, hyper-parameters $\lambda_1,\lambda_2, \lambda_3$.\\
\textbf{Output:} clustering assignment.
\begin{algorithmic}[1] 
\FOR{$t\leftarrow$ 1 to $epoch$ } 
\FOR{b $\leftarrow$ 1 to $ \lceil N/batchsize \rceil $}
\STATE conduct $K$-means on each view to obtain the cluster assignment 
\STATE compute the silhouette coefficient $\{sils^{v}\}_{v=1}^V$ by Eq. (\ref{eq7})
\STATE // For RG-UMC
\STATE compute the most reliable view index $r$ by Eq. (\ref{eq6})
\STATE update the autoencoder parameters $\{F\}_{v=1}^V$ and $\{G\}_{v=1}^V$ by minimizing the total loss $L_{RG}$ in Eq. (\ref{eq13}).
\STATE // For RGs-UMC
\STATE compute the weight matrix $w_{vr}$ by Eq. (\ref{eq16})
\STATE update the autoencoder parameters $\{F\}_{v=1}^V$ and $\{G\}_{v=1}^V$ by minimizing the total loss $L_{RGs}$ in Eq. (\ref{eq14}).
\ENDFOR
\ENDFOR
% \STATE Feed the unpaired dataset into network to obtain the latent representation $\{\boldsymbol{Z}^v\}_{v=1}^V$ for all views
\STATE Utilize the unpaired dataset as input to the network and obtain the latent representation ${\boldsymbol{Z}^v}_{v=1}^V$ for all views.
\STATE Perform $K$-means clustering on $\boldsymbol{Z}$ to derive the ultimate clustering assignment, where $\boldsymbol{Z}=[\boldsymbol{Z}^1; \boldsymbol{Z}^2; \ldots; \boldsymbol{Z}^V]$.
% \STATE Conduct $K$-means on $\boldsymbol{Z}$ to obtain the final clustering assignment, where $\boldsymbol{Z}=[\boldsymbol{Z}^1; \boldsymbol{Z}^2; \ldots; \boldsymbol{Z}^V]$.
\end{algorithmic}
\end{algorithm}

\section{Experiments}\label{experiments}
%\re{Add several sentences to introduce the section.}
% We first introduce the datasets, experimental setting, and compared methods. Then, we compare the clustering performance of RG-UMC and RGs-UMC with the compared methods and followed by visualization, ablation study, and parameter analysis. {Specifically, we conduct an experiment to validate the effect of a single reliable view and multiple reliable views guidance dynamically over a fixed reliable view.} 

% In this section, we compare our models RG-UMC and RGs-UMC with the sixteen comparison methods and followed by ablation studies, visualization, and other experiments to validate the effectiveness of our reliable view guidance strategies. 

{In this section, we evaluate the performance of our models, RG-UMC and RGs-UMC, against sixteen comparison methods. Additionally, we conduct ablation studies, visualization, and other experiments to confirm the effectiveness of our reliable view guidance strategies.}

For convenience, some information including experimental setting, network architectures, 5 benchmark multi-view datasets (\emph{Digit}\footnotemark[1], \emph{Scene-15}\footnotemark[2], \emph{Caltech101-20}\footnotemark[2], \emph{Flower17}\footnotemark[3] and \emph{Reuters}\footnotemark[4]), {16} state-of-art compared methods, and parameter analysis are provided in the supplementary.
Specifically, {16} state-of-art methods include five complete multi-view clustering methods (OPMC \cite{liu2021onelarge}, OP-LFMVC \cite{liu2021one}, DUA-Nets \cite{2022Uncertainty}, {RMVC}\cite{tao2018reliable}, and {DSMVC} \cite{tang2022deep}), eight incomplete multi-view clustering methods (DAIMC \cite{2018Doubly}, UEAF \cite{Jie2019Unified}, IMSC-AGL \cite{2020Jie}, OMVC \cite{Shao2017Online}, OPIMC \cite{2019One}, MvCLN \cite{yang2021MvCLN}, Completer \cite{lin2021completer}, and {T-UMC}\cite{lin2022tensor}), and three unpaired multi-view clustering methods (IUMC-CA \cite{10149819}, IUMC-CY \cite{10149819} and scl-UMC \cite{scl-UMC}).

\footnotetext[1]{http://archive.ics.uci.edu/dataset/72/multiple+features}
\footnotetext[2]{https://github.com/XLearning-SCU/2021-CVPR-Completer/tree/main/data}
\footnotetext[3]{http://www.robots.ox.ac.uk/vgg/data/flowers/17/index.html}
\footnotetext[4]{http://archive.ics.uci.edu/ml/datasets/Reuters+RCV1+RCV2+Multilingual\\\%2C+Multiview+Text+Categorization+Test+collection}

\subsection{Performance Comparison and Analysis}
% \textbf{(1) Comparison of clustering performance with two views.} TABLE \ref{table4_comparison} reports the clustering performance on five datasets, where the top two results in each column are highlighted in bold and underlined, respectively. In particular, for the \emph{Digit}, \emph{Flower17}, and \emph{Reuters} datasets, the first two views are used as the two-view dataset. For the other two datasets, the views are selected following Completer \cite{lin2021completer} and DCP \cite{lin2022dual}. From TABLE \ref{table4_comparison}, we make the following observations: 
\textbf{(1) Evaluating clustering performance with two views.} {TABLE \ref{table4_comparison} presents the clustering results across five datasets. %The top two results in each column are highlighted in bold and underlined, respectively. 
{The top two results in each column are emphasized by being highlighted in bold and underlined, respectively.} Specifically, the \emph{Digit}, \emph{Flower17}, and \emph{Reuters} datasets utilize the first two views as the two-view dataset. As for the remaining two datasets, the two views are selected according to Completer \cite{lin2021completer} and DCP \cite{lin2022dual}. From TABLE \ref{table4_comparison}, we draw the conclusions:}
%Specifically, \emph{Digit} and \emph{Flower17} datasets use the first two views as the two views dataset, while the other three datasets choose views the same with Completer \cite{lin2021completer} and DCP \cite{lin2022dual}. 
%Specifically, \emph{Caltech101-20} choose the last two feature HOG feature (1984-D) and GIST feature (512-D) as two views. \emph{Scene-15} choose PHOG (20-D) and GIST (59-D) features as two different views. %\emph{Reuters} choose the first two languages, English-English (21531-D) and French-English (24893-D), as two views in the experiment. 

1) Compared with the first thirteen clustering methods (complete and incomplete multi-view clustering methods), RG-UMC and RGs-UMC significantly outperform most of the compared methods. In UMC, where paired samples between views are absent, existing methods often struggle to capture cross-view relationships and may deteriorate into models for single-view. Fortunately, our methods leverage reliable view guidance to construct the relationship and achieve a consistent cluster structure between views, resulting in superior results.
% In the case of UMC, where there are no matched samples between views, these compared methods struggle to capture cross-view relationships and tend to degenerate into single-view models. In contrast, our methods learn consistent cluster structure by reliable view guidance, thus achieving better performance. 
2) Compared with the three UMC methods, our models exhibit significant improvements across nearly all evaluation metrics, indicating that the guidance of reliable views has positive effects on other views. {Specifically, RG-UMC shows an average improvement of 23.46\%, 32.86\%, and 27.88\% in NMI, ACC, and F-score, respectively, compared to these methods.} Similarly, RGs-UMC exhibits a higher average improvement of 23.78\%, 33.69\%, and 30.18\% in NMI, ACC, and F-score, respectively.
% Specifically, RG-UMC achieves an average improvement of 23.46\% in NMI, 32.86\% in ACC, and 27.88\% in F-score compared to them. Similarly, RGs-UMC achieves an even higher average improvement of 23.78\%, 33.69\%, and 30.18\% in NMI, ACC, and F-score respectively.
%Specifically, RG-UMC achieves an average improvement of 24.14\% in NMI, 37.49\% in ACC, and 33.05\% in F-score compared to IUMC-CA and IUMC-CY. Similar, RGs-UMC achieves an even higher average improvement of 29.42\%, 41.36\%, and 35.47\% in NMI, ACC, and F-score respectively. % and the effectiveness of each module on reliable KL divergence will be further analyzed in the ablation study. % Furthermore, the latent representation learned by the autoencoder is better than that of Matrix decomposition, which validates the power of deep learning.
% 3) Compare RG-UMC with RGs-UMC, it is worth mentioning that RGs-UMC, which comprehensively explores inter-view relationships, outperforms RG-UMC in terms of performance.% and facilitates clustering formation. 
3) Compare RG-UMC with RGs-UMC, the performance is comparable on the two views dataset. The reason may be that these two methods are nearly equivalent when only one reliable view is selected. %, which can be analyzed by Eq. (\ref{eq6}) and Eq. (\ref{eq16}). % Specifically, from Eq. (\ref{eq6}) and Eq. (\ref{eq16}), we observed that the two equations are nearly equivalent when only one reliable view is selected.

\begin{table*}%[t]
\centering
% \vspace{-1mm}
% \renewcommand{\arraystretch}{1.1}
% \setlength{\tabcolsep}{5.5pt}
\renewcommand{\arraystretch}{1.1}
\setlength{\tabcolsep}{5.5pt}
% \caption{Comparison of related multi-view clustering methods for unpaired \textbf{two-view} clustering. The table presents the results of \emph{NMI}, accuracy (\emph{ACC}), and F1-score (\emph{F1}) in percentage, with the top two results in each column highlighted in bold and underlined, respectively.}\label{table4_comparison}
\caption{Comparison of related multi-view clustering methods for UMC with \textbf{two-view}. The results are presented in percentage, with the top two results in each column highlighted in bold and underlined, respectively.}\label{table4_comparison}
% \vspace{-0.3cm}
\begin{tabular}{l|ccc|ccc|ccc|ccc|ccc}
%\toprule[1pt]
\hline\hline
\multirow{2}{*}{\textbf{Methods}}& \multicolumn{3}{c}{\emph{Digit}} \vline &\multicolumn{3}{c}{\emph{Scene-15}} \vline &\multicolumn{3}{c}{\emph{Caltech101-20}} \vline
&\multicolumn{3}{c}{\emph{Flower17}}\vline
&\multicolumn{3}{c}{\emph{Reuters}}\\ \cline{2-16}
%\cline{2-4}\cline{6-8}\cline{10-12}\cline{14-16} 
%&&\multicolumn{3}{c}{\emph{Office31}}\\\cline{2-5}\cline{6-9}\cline{10-13}\cline{14-16}
&\emph{NMI} &\emph{ACC} &\emph{F1} &\emph{NMI} &\emph{ACC} &\emph{F1} &\emph{NMI} &\emph{ACC} &\emph{F1} &\emph{NMI} &\emph{ACC} &\emph{F1}&\emph{NMI} &\emph{ACC} &\emph{F1}\\\hline
\textbf{OPMC} \cite{liu2021onelarge}&44.12	&51.75	&32.02	&30.43	&27.29	&19.76	&27.59	&28.12	&24.71	&26.19	&24.56	&14.91 &20.17	&40.32	&{36.73}\\ %\hline
\textbf{OP-LFMVC} \cite{liu2021one}  &39.04	&46.40	&35.10	&15.60	&20.45	&13.28	&21.61	&22.09	&17.46	&20.81	&23.31	&13.09 &- &- &-\\ %\hline
% \textbf{MFLVC} \cite{xu2022multi} &14.66 &14.10	&10.01  & 5.19	&5.93	&3.48 & 8.56	&8.01	&2.57 &8.76	&7.50	&8.62  &20.19	&20.18	&17.13\\
\textbf{DUA-Nets}\cite{2022Uncertainty}&42.08 &38.07 &33.98 &30.12	&27.80	&21.23 &32.40	&29.26	&25.42 & 28.34 &24.44	&18.61 &4.28 &26.23	&24.17\\
 %&1.06 &19.34	&19.36
\textbf{RMVC} \cite{tao2018reliable} &46.75  &44.05 &36.56 &36.34 &30.52 &21.92  &34.37	&24.43   &22.11 &33.70	&28.24	&18.83 &- &- &-\\
\textbf{DSMVC} \cite{tang2022deep} &49.84 &41.90	&8.39 &27.00	&25.22	&6.03 &32.14& 22.34&5.29 &26.69 &23.46 &4.14 &10.86 &	30.39 &14.40\\

\textbf{DAIMC} \cite{2018Doubly} &39.37	&38.50	&30.88	&19.46&	22.79	&16.20	&20.16	&24.81	&22.88	&21.87	&21.03	&15.12 &- &- &- \\ %\hline
\textbf{UEAF} \cite{Jie2019Unified} &30.90	&26.77	&19.32	&22.16	&20.14	&13.19	&25.44	&22.19 	&32.54	&12.72	&9.25	&6.37 &- &- &-\\ %\hline
\textbf{IMSC-AGL} \cite{2020Jie}&48.99 &45.40 &37.09&27.61	&27.98	&18.33 &30.04	&24.06	&20.38 &26.99 &24.71 &15.31  &- &- &-\\
\textbf{OMVC} \cite{Shao2017Online} &31.69	&34.87	&20.64 & 0.68& 9.16&12.94 &26.11&26.78	&22.62 &26.08	&24.52	&14.47 &-&-&-
\\ %\hline
\textbf{OPIMC} \cite{2019One}&41.49 &44.60 &37.01 &20.32&23.52&15.30 &22.38&24.48&21.97 &19.90&21.84&12.22 &8.99	&27.53	&24.36\\ 
\textbf{MvCLN} \cite{yang2021MvCLN} &41.96	&45.10	&52.41 & 21.50	&21.43	&22.47 &31.21	&28.04	&17.64 &19.57	&17.65	&18.43 &20.18	&18.97	&19.72\\
\textbf{Completer}\cite{lin2021completer}&41.35 &39.66	&40.24	&23.24	&21.86	&19.59	&34.29	&26.37	&22.61	&22.65	&20.73	&19.83 &1.28	&23.84	&19.67\\
% reuters &0.59	&20.90	&17.34
\textbf{T-UMC}\cite{lin2022tensor} &46.15	&49.85	&49.26 &40.75	&39.20	&30.10 &38.84	&35.88	&23.54 &22.44	&22.65 &16.51 &-&-&-\\
\textbf{IUMC-CA}\cite{10149819}&46.20	&27.75	&29.82	&{40.50}	&30.08	&27.39	&35.91	&26.40	&19.24	&\bf{43.31}	&{31.84}	&25.64 &- &- &-\\ 
\textbf{IUMC-CY}\cite{10149819}&49.17	&45.55	&37.72	&39.67	&18.68	&23.87	&32.22	&28.08	&{35.06}	&39.52	&18.82	&21.01 &11.34	&27.50	&27.36\\
\textbf{scl-UMC}\cite{scl-UMC}&{57.00}&{60.85}&{61.35}	&{45.26}	&{36.59}	&{35.35} &{38.60}	&{41.52}	&{36.84}&{39.98}	&{32.21}	&{35.06} & {39.50}	&{49.86}	&\underline{41.12}\\
\textbf{RG-UMC}&\underline{92.38}&\underline{96.40}&\underline{96.42}	&\underline{46.79}	&\underline{49.53}	&\underline{47.40} 
&\bf{78.64}	&\bf{73.89}	&\bf{55.46}
&\underline{42.01}	&\bf{43.75}	&\bf{42.80}	
&\underline{53.56}	&\underline{60.53}	&{39.13}\\

\textbf{RGs-UMC}&\bf{93.93}&\bf{97.20}&\bf{97.21}    &\bf{48.57}&\bf{52.19}&\bf{50.70}    &\underline{76.37}&\underline{71.17}&\underline{54.50}  &{41.35}&\underline{41.62}&\underline{42.30}  &\bf{55.82}&\bf{71.96}&\bf{56.18}\\
\hline\hline
\end{tabular}
% \vspace{-0.1cm}
\end{table*}

\textbf{(2) Evaluating clustering performance with multiple views.}
{We compare our models with other comparison methods on multi-view datasets containing all views.} %In the unpaired multi-view scenario, each sample is observed in only a view, and the observation ratio for each view is $1/V$. 
% \re{In UMC, each sample is only observed in one view, with the observation ratio for each view being $1/V$.} 
{In UMC, each sample is observed in only one view, resulting in an observation ratio of $1/V$ for each view.} The number of observed samples per view decreases as the number of views increases, posing a challenge to effectively address UMC. %As the number of views increases, the number of observation samples per view decreases, making it challenging to learn unpaired multi-view clustering effectively. % For this experiment, we compare \xlk{twelve} comparison methods except for Completer and MvCLN, because they are designed for two-view clustering. Specifically, TABLE \ref{table5_multi_view_morethantwo} reports the clustering performance on five datasets, where the top two results in each column are highlighted in bold and underlined, respectively. From the TABLE \ref{table5_multi_view_morethantwo}, we have following observations:
% For this experiment, we compare {fourteen} comparison methods, excluding Completer and MvCLN, as they are designed for two-view clustering. The clustering performance on five datasets is reported in TABLE \ref{table5_multi_view_morethantwo}. From TABLE \ref{table5_multi_view_morethantwo}, we make the following observations: 
% \xlk{For this experiment, we exclude Completer and MvCLN, which are designed for two-view clustering \cite{scl-UMC}, and compare our methods with other fourteen approaches. The clustering performance across five datasets is presented in TABLE \ref{table5_multi_view_morethantwo}. From the results, we draw the following observations: }
For this experiment, we omit Completer and MvCLN, which are tailored for two-view clustering \cite{scl-UMC}, and evaluate our methods against fourteen other approaches. The clustering performance across five datasets is presented in TABLE \ref{table5_multi_view_morethantwo}. From the results, we observed:

1) Compared with both complete and incomplete multi-view clustering methods, our models, RG-UMC and RGs-UMC, outperform most comparison methods, demonstrating their effectiveness in multi-view clustering with multiple views. 2) Compared with three unpaired multi-view clustering methods, our models have better performance on most evaluation metrics, indicating that the guidance of reliable views has positive effects on other views. {In particular, RG-UMC exhibits an average increase of 1.85\% in NMI, 18.85\% in ACC, and 10.03\% in F-score compared to them. Similarly, RGs-UMC demonstrates an even greater average improvement of 5.17\%, 22.62\%, and 12.80\% in NMI, ACC, and F-score, respectively.}
{Although IUMC-CY achieves better results in NMI than our methods on the \emph{Scene-15 (3 views)} dataset, its accuracy and F-score are lower by 22.92\% and 12.76\%, respectively. }Additionally, we introduce another indicator, precision, which is 23.79\% for IUMC-CY and 47.84\% for RGs-UMC. With the additional indicator, our method performs relatively well across all four indicators. 3) Compared with TABLE \ref{table4_comparison}, the clustering performances of the comparison methods are generally better in two views rather than multiple views. The reason is that the distribution difference between observed samples increases with the number of views increases. This leads to greater difficulty in aligning without pairing, resulting in decreased performance for the comparison methods. However, our RG-UMC and RGs-UMC exhibit a relatively small decrease in performance, indicating the effectiveness of reliable view guidance. %models show only minimal performance decline, % or even improvement in some datasets (\eg flower17), % Our method exhibits a relatively small decrease in performance.
% This could be due to the close distance between clusters, leading to confusion in assigning boundary samples and resulting in reduced accuracy and F-score. 

\begin{table*}%[t]
\centering
% \vspace{-1mm}
\renewcommand{\arraystretch}{1.1}
\setlength{\tabcolsep}{5.5pt}
% \caption{Comparison of related multi-view clustering methods for unpaired \textbf{multi-view} clustering. The cells represent the results of \emph{NMI}, accuracy (\emph{ACC}), and F1-score (\emph{F1}) in percentage, with the top two results in each column highlighted in bold and underlined, respectively.}\label{table5_multi_view_morethantwo}
\caption{Comparison of related multi-view clustering methods for UMC with \textbf{multiple views}. }\label{table5_multi_view_morethantwo}
% \vspace{-0.3cm} % The results are presented in percentage, with the top two results in each column highlighted in bold and underlined, respectively.
\begin{tabular}{l|ccc|ccc|ccc|ccc|ccc}
%\toprule[1pt]
\hline\hline
\multirow{2}{*}{\textbf{Methods}}& \multicolumn{3}{c}{\emph{Digit (6views)}} \vline &\multicolumn{3}{c}{\emph{Scene-15 (3views)}} \vline &\multicolumn{3}{c}{\emph{Caltech101-20 (6views)}} \vline
&\multicolumn{3}{c}{\emph{Flower17 (7views)}}\vline&\multicolumn{3}{c}{\emph{Reuters (5views)}}\\ \cline{2-16}
%\cline{2-4}\cline{6-8}\cline{10-12}\cline{14-16} 
%&&\multicolumn{3}{c}{\emph{Office31}}\\\cline{2-5}\cline{6-9}\cline{10-13}\cline{14-16}
&\emph{NMI} &\emph{ACC} &\emph{F1} &\emph{NMI} &\emph{ACC} &\emph{F1} &\emph{NMI} &\emph{ACC} &\emph{F1} &\emph{NMI} &\emph{ACC} &\emph{F1}
&\emph{NMI} &\emph{ACC} &\emph{F1}\\\hline

\textbf{OPMC}\cite{liu2021onelarge}  &12.31&18.33&17.42   &17.23&18.94&14.01  &12.42&14.30&14.69  &13.52&13.38&10.14  &11.11&19.48&20.41 \\
\textbf{OP-LFMVC}\cite{liu2021one}
&22.54&11.57&19.90  &9.73&17.02&11.02   &13.38&17.34&13.56  &9.71&14.88&8.02    &-&-&-  \\ 
\textbf{DAU-Nets}\cite{2022Uncertainty}&17.10&19.40&16.82    &22.62&21.52&16.30  &24.50&15.34&17.90  &14.84&13.81&10.77  &2.20	&25.06	&22.98 \\
\textbf{RMVC}\cite{tao2018reliable} &22.31	&26.25	&17.88    &28.84 &24.68 &	18.03 &21.72 &18.90 &17.34 &19.08 &17.28	&11.24 &- &- &-\\
\textbf{DSMVC}\cite{tang2022deep}& 13.95	&16.50	&7.80 & 19.86	&19.38 &10.09 &19.44 &14.71 &2.60 &13.55 & 12.50	&4.57 &1.53 &19.06 &15.12\\

\textbf{DAIMC}  \cite{2018Doubly}
&18.17&25.68&17.96  &12.49&19.46&18.68  &12.49&19.46&18.68  &8.56&12.75&9.83     &-&-&- \\
\textbf{UEAF}\cite{Jie2019Unified}
&10.32&19.25&16.63  &11.59&15.65&12.56  &9.28&27.03&24.26   &3.91&7.54&10.74    &-&-&- \\
\textbf{IMSC-AGL}\cite{2020Jie}
&24.01&17.61&17.38  &16.45&20.88&12.99  &17.30&15.81&12.97  &12.66&14.93&8.69    &-&-&- \\ %\hline
\textbf{OMVC}\cite{Shao2017Online}
&12.02&20.03&16.70  &11.53&16.09&11.74  &18.29&19.26&17.31  &12.68&14.68&9.13    &-&-&- \\
\textbf{OPIMC} \cite{2019One}  &20.13&27.74&19.66  &15.48&20.33&13.35  &17.10&20.18&16.71  &11.76&16.39&8.87   &10.92&27.46&26.13 \\
% \textbf{MFLVC}\cite{xu2022multi}& 10.03   &10.15	&10.08 &33.27&5.95&3.68  &9.48	&11.36	&4.64    &7.65&7.13&3.12   &28.76 &24.66 &20.75\\
\textbf{T-UMC}\cite{lin2022tensor} &9.09	&17.70	&15.06
&26.51	&27.65	&22.91 &11.95	&14.08	&9.01 &11.66	&15.74 &13.57 &-&-&-\\

\textbf{IUMC-CA}\cite{10149819}
&75.16&54.53&60.82    &\underline{50.25}&{34.98}&34.47 &58.31&33.42&33.24  &\textbf{66.13}&{48.52}&{44.19}   &-&-&- \\ 
\textbf{IUMC-CY}\cite{10149819}
&{74.64}&{60.72}&{61.17}   &\textbf{56.75}&30.58&{35.34} &{61.48}&{46.63}&\bf{45.06}   &\underline{63.23}&{49.21}&40.19  &10.07&22.99&21.48 \\ 
\textbf{scl-UMC}\cite{scl-UMC}&{72.75}	&{60.00}	&{50.97} &49.62	&39.15	&33.51    &59.63	&57.94	&{60.93}    &{65.65}	&49.93	&47.80  &30.73 &52.83&31.32 \\ 
\textbf{RG-UMC} &\underline{77.37} &\underline{84.85}&\underline{84.65}  &{49.23}	&\textbf{53.82}	&\textbf{49.82}  &\underline{65.81}&\underline{67.05}&{38.92}      &58.57&\underline{50.04}&\underline{47.81}	&\underline{30.93}&\textbf{59.78}&\textbf{34.54} \\
% digit-5views &\underline{87.42} &\underline{93.10}&\underline{93.21} 
% scene15-3views  &{49.79}	&\underline{52.79}	&\underline{47.72}

% \textbf{UKL-UMC}&97.86	&98.80	&98.80    &42.01	&48.02	&46.04    &65.42	&{73.35}	&{41.67}    &55.02	&49.50	&49.35    &22.03	&46.16	&31.52\\
% \textbf{NKL-UMC}&97.79	&98.95	&98.95   &45.10	&51.47	&46.66    &61.09	&62.25	&33.74    &53.41	&48.05	&45.61     &27.30	&54.71	&32.12\\
\textbf{RGs-UMC} &\textbf{86.19}&\textbf{92.83}&\textbf{92.77}       &47.40&\underline{53.50}&\underline{48.11}    &\textbf{70.88}&\textbf{75.28}&\underline{41.50}     &{60.64}&\bf{52.02}&\bf{51.68} &\textbf{32.26}&\underline{57.24}&\underline{33.35}\\
% digit-5views &\textbf{97.95}&\textbf{99.05}&\textbf{99.05}  

\hline\hline
\end{tabular}
% \vspace{-0.1cm}
\end{table*}

\subsection{Ablation Study and Parameter Analysis}

{\textbf{Ablation study of different losses on RG-UMC and RGs-UMC models.}} % To further investigate the performance of RG-UMC and RGs-UMC, we conduct the ablation study on \emph{Digit} with the first two views. Specifically, the Orthogonal constraints on $\{\boldsymbol{Z}^v\}_{v=1}^V$ and the compactness learning module in RG-UMC and RGs-UMC model are abbreviated as $Orth$ and $\mathcal{C}$, respectively. The KL divergence learning module in RG-UMC and RGs-UMC models abbreviate as $KL$ and $KLs$, respectively. And their different combinations are evaluated in TABLE \ref{ablationstudy-KL-UCM}.
To further analyze the modules in RG-UMC and RGs-UMC, we conducted ablation studies on \emph{Digit} dataset with two views. We evaluated different combinations of the orthogonal constraint ($Orth$), the compactness learning module ($\mathcal{C}$), and the alignment module ($KL$ for RG-UMC and $KLs$ for RGs-UMC). The results are presented in TABLE \ref{ablationstudy-KL-UCM}.
\begin{table}
% \vspace{-0.1cm}
\centering
\renewcommand{\arraystretch}{1.05}
\setlength{\tabcolsep}{7pt}
\caption{Ablation study of different losses on RG-UMC and RGs-UMC models, conducted on the \emph{Digit} dataset with two views.}\label{ablationstudy-KL-UCM}
% \vspace{-0.3cm}
\begin{tabular}{c|cccc|ccc}
\hline\hline
Line &\emph{Orth} &$\mathcal{C}$ &\emph{KL} &\emph{KLs}& \emph{NMI} & \emph{ACC} & \emph{F1}\\

\hline
1&\quad&  \quad & \quad & \quad& 41.77& 42.25& 45.88\\%\hline
2&\checkmark&  \quad & \quad & \quad& 41.72 &44.35 &48.48\\%\hline
3&\quad &\checkmark& \quad  & \quad&44.45 &44.20 & 46.34\\
4&\checkmark& \checkmark &\quad	& \quad&46.28&46.65&48.86\\\hline
5&\quad & \quad & \checkmark &- &82.83 & 91.40 &91.43\\
6&\checkmark& \quad &\checkmark  &- &91.80 &95.85&95.88\\%\hline
7&\quad &\checkmark &\checkmark	&- &91.97&96.25&96.27\\%\hline
8&\checkmark& \checkmark& \checkmark&- & \textbf{92.38}	&\textbf{96.40}&\textbf{96.42}\\\hline
9&\quad & \quad &-& \checkmark &92.93	&96.60	&96.62\\
10&\checkmark& \quad &-&\checkmark  &93.67	&97.05	&97.07\\%\hline
11&\quad &\checkmark &-&\checkmark	&92.65	&96.45	&96.47\\%\hline
12&\checkmark& \checkmark&- &\checkmark& \textbf{93.93}	&\textbf{97.20}&\textbf{97.21}\\
% \bottomrule[1pt]%
\hline\hline
\end{tabular}
\vspace{-0.2cm}
\end{table}

%As observed, the combination of all modules performs the best (Line 8 and line 12), which is our method KL-UMC and RGs-UMC. Specifically, comparing Lines 2-4 with Line 1, the addition of reliable KL divergence module aligns the subspace representations, which improves the model performance effectively. While the addition of the Orthogonal constraints module and compactness module has only a small effect on the clustering result. That means, the main task in unpaired multi-view clustering is learning the consistency information among views. In Lines 5-7, with the help of other terms, the clustering performance improved further. For example, combined with $Orth$ and $\mathcal{C}$ modules, $KL$ module has better performance than single one for the latent representation. %$Orth$ module aims \cite{chen2022adaptively} to avoid pushing the integral space arbitrarily, meanwhile attempts to attain mutual independence between different sample latent representation. The orthogonal constraint restrict the latent representation to be more discriminative \cite{chen2022efficient}.
As observed, the combination of all modules performs the best (Line 8 and Line 12), which is our method RG-UMC and RGs-UMC. Lines 1-8 are the ablation study of RG-UMC, while Lines 1-4 and Lines 9-12 are the ablation study of RGs-UMC. Comparing Lines 2-4 with Line 1, the $Orth$ and $\mathcal{C}$ modules have positive effects on clustering. Specifically, comparing Lines 2,3,5,9 with Line 1, adding the $KL$ and $KLs$ modules significantly improves performance by guiding the learning of subspace representations, while the impact of the $Orth$ and $\mathcal{C}$ modules is minor. That further validates the main task in UMC is to learn the consistency and alignment among views.
%Specifically, comparing Lines 2,3,5,9 with Line 1, the addition of the $KL$ module and $KLs$ module aligns the subspace representations, which improves the model performance effectively. While the addition of $Orth$ and $\mathcal{C}$ modules have only a small effect on the clustering result. That means the main task in unpaired multi-view clustering is learning the information of consistency and alignment among views.
Lines 5-8 outperformed Lines 1-4, demonstrating the effectiveness of the $KL$ module. Similarly, comparing Lines 9-12 with Lines 1-4, with the help of the $KLs$ module, the clustering performance improved further. Finally, comparing Lines 5-8 and Lines 9-12, RGs-UMC shows slightly better performance than RG-UMC on all evaluation metrics.
%That means the main task in unpaired multi-view clustering is learning the consistency of information among views. 
% \textbf{Effect of adaptive reliable view selection in RG-UMC and RGs-UMC models.}
% To investigate the influence of adaptive reliable view selection in RG-UMC and RGs-UMC models, we take Scene15 with 3 views for example and show the chosen reliable view at each epoch. Specifically, we compare our models with T-UMC and the result is shown in Fig. \ref{reliable_idx_TUMC_RGUMC_RGsUMC}. In T-UMC, a fixed single view is chosen as the reliable view. In RG-UMC, a single view is dynamically selected as the reliable view during optimization. In RGs-UMC, multiple views are designated as reliable views, leading to a larger number of reliable views. This increased number of reliable view guidance helps accelerate the learning process. %As Fig. \ref{reliable_idx_TUMC_RGUMC_RGsUMC} (a) shows, T-UMC has the fixed and single view (\eg view 1) as reliable view. In Fig. \ref{reliable_idx_TUMC_RGUMC_RGsUMC} (b), RG-UMC has single view as reliable view, where the view is adaptive changed during optimization, at each epoch. In Fig. \ref{reliable_idx_TUMC_RGUMC_RGsUMC} (c), RGs-UMC has multiple views as reliable view, and the number of reliable view is expanded largely. The increase number of reliable view guidance accelerates the learning process.

\begin{figure}%[!t]
\centering
\includegraphics[width=1.0 \columnwidth]{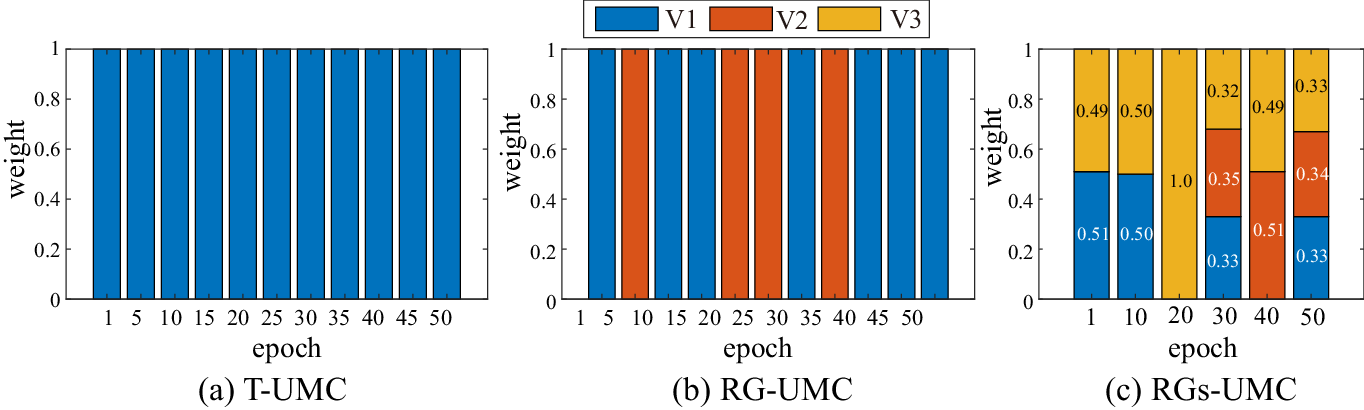}
\vspace{-0.4cm}
\caption{Comparison of adaptive reliable view selection in RG-UMC and RGs-UMC models with T-UMC.}
\label{reliable_idx_TUMC_RGUMC_RGsUMC}
\vspace{-0.2cm}
\end{figure}

\textbf{Effect of reliable view selection adaptively in RG-UMC and RGs-UMC models.}
For the limitations of a fixed and single reliable view in T-UMC, we investigate an adaptive selection of reliable views in RG-UMC and RGs-UMC by the \emph{Scene-15} dataset with 3 views. As shown in Fig. \ref{reliable_idx_TUMC_RGUMC_RGsUMC}, the weights of these reliable views sum up to 1. We show the weight on each view every five (in RG-UMC) or ten (in RGs-UMC) epochs. In T-UMC, only view 1 is consistently chosen as the reliable view. In contrast, RG-UMC dynamically alternates between view 1 and view 2 as the reliable view. RGs-UMC takes it a step further by designating view 3 as a reliable view during the optimization process. The number of reliable views increased enhancing the complementary between views.
% The results are shown in Fig. \ref{reliable_idx_TUMC_RGUMC_RGsUMC}, where the view selected denotes solid points. In T-UMC, a single view is pointed as the reliable view, which remains fixed throughout the optimization process. In contrast, RG-UMC dynamically selects a single view as the reliable view during optimization. RGs-UMC goes a step further by designating multiple views as reliable views, resulting in a larger number of reliable views. This increased number of reliable view guidance facilitates faster learning and enhances the complementary between views.
%For the limitations may exit in T-UMC with a fixed and single reliable view: i) it restricts the guidance of the more reliable view that may emerge during the optimization process. ii) it overlooks the guiding effect of other suboptimal views. We validate the effect of adaptive reliable view selection in RG-UMC and RGs-UMC models using the Scene15 dataset with 3 views. The result is shown in Fig. \ref{reliable_idx_TUMC_RGUMC_RGsUMC}. In T-UMC, a fixed single view is chosen as the reliable view. In RG-UMC, a single view is dynamically selected as the reliable view during optimization. In RGs-UMC, multiple views are designated as reliable views, leading to a larger number of reliable views. This increased number of reliable view guidance helps accelerate the learning process. 

\textbf{Influence of reliable strategy in RG-UMC model.}
%To validate the reliable performance of the reliable KL module in RG-UMC model, we design three replacements (appoint strategy, random strategy, reliable strategy ($V-1$)) for the reliable strategy ($V$). The appoint strategy means appointing a view (\eg the first view) as the aligned view. The random strategy means choosing to align the view randomly as training goes on. The reliable strategy has two modes, the reliable view align with other views including itself ($V, V-1$) or not.`reliable’
To evaluate the reliable view guidance in RG-UMC, we employed three alternative strategies: `appoint’, `random’, and `reliable ($V-1$)’. `Appoint’ entails selecting a specific view (\eg view 1) as a reliable view throughout training, `random’ denotes choosing a view randomly at each iteration, and `reliable ($V-1$)’ means aligning the most reliable view with other views, excluding itself. By comparing clustering performance under these strategies, we validate reliable view guidance effectiveness in RG-UMC. % To assess the effectiveness of reliable view guidance in the RG-UMC model, we used three alternative strategies: appoint, random, and reliable ($V-1$). The appoint strategy selects a specific view (\eg the first view) as the reliable view to guide throughout training, while the random strategy randomly selects a view as the reliable view to align other views at each iteration. In the reliable ($V-1$) strategy, the reliable view aligns with the other $V-1$ views, excluding itself. By comparing the clustering performance under these strategies, we can validate the effectiveness of the reliable view guidance in the RG-UMC model. %we can evaluate the impact of reliable view alignment and validate the effectiveness of the reliable view guidance in the RG-UMC model. % The experiment was conducted on the digit data set with five views and the flower17 data set with seven views. We evaluated the clustering performance of each view using different strategies: appoint, random, reliable ($V-1$), and reliable ($V$). 
The experiment is conducted on the five data sets with all views in Table \ref{reliable-KL-UMC}. Specifically, the `all-view’ and other lines (\eg `v1’, `v2’, \ldots, `v7’) are the clustering results on the dataset with all views and single views, respectively. Additionally, the `Original’ column is the clustering result with $K$-means on the Original data. The `appoint’, `random’, and `reliable ($V-1$)’ columns are the strategies mentioned above and the last column is our model RG-UMC. 

From TABLE \ref{reliable-KL-UMC}, {several conclusions can be drawn.} The `all-view’ line shows that the reliable strategy yields the best performance, while the performance of the `appoint’ and `random’ strategies is unstable, which is related to the randomness of reliable view selection. Additionally, the reliable strategy with all views performs slightly better than the strategy with `reliable ($V-1$)’ views, indicating the benefit of aligning the reliable view with itself. Furthermore, comparing the single view performance on original data sets and latent representation with four strategies, we observed that the single view performance has improved through all variants of the reliable alignment module via KL divergence.
% From the 'all-view' line of TABLE \ref{reliabel-KL-UMC}, we observed that the strategy of choosing a reliable view has the best performance, the selecting randomly strategy has the second performance, and specifying a view has the worst performance. Then, we validate the efficiency of the reliable strategy. Besides, the reliable strategy with V-views is slightly better than the reliable strategy with (V-1)-views, which means that aligning with the self is effective.
% The 'all-view' line of TABLE \ref{reliable-KL-UMC} shows that the reliable strategy yields the best performance, followed by the random strategy, and the appoint strategy performs the worst. 
% \begin{table}
% \vspace{-0.1cm}
% \centering
% %\small
% \renewcommand{\arraystretch}{1.0}
% \setlength{\tabcolsep}{7pt}
% \caption{Influence of appoint view, random view, reliable view of KL-UMC on \emph{Digit} with five views}\label{reliabel-KL-UMC}
% \vspace{-0.3cm}
% \begin{tabular}{c|c|ccc}
% %\toprule[1pt]
% \hline\hline
% Dataset &\emph{description} & \emph{NMI} & \emph{ACC} & \emph{F1}\\
% \hline
% \multirow{4}{*}{\textbf{digit-5view}}
% & appoint & 89.35	& 91.75	&91.95\\%\hline
% & random & 91.99	&94.8	&94.83\\%\hline
% & reliable(V-1) &93.75	&96.25	&96.25\\
% & reliable(V) &94.05	&96.45	&96.45\\
% \multirow{4}{*}{\textbf{flower17-7view}}
% & appoint &16.25	&15.13	&15.51\\%\hline
% & random &16.28	&16.88	&16.12\\%\hline
% & reliable(V-1) &18.08	&16.81	&16.06\\
% & reliable(V) &18.57	&16.42	&16.26\\
% \hline\hline
% \end{tabular}
% \vspace{-0.4cm}
% \end{table}

\begin{table*}%[t]
\centering
%\vspace{-1mm}
\renewcommand{\arraystretch}{1.1}
\setlength{\tabcolsep}{4.5pt}
\caption{Influence of original view, appoint view, random view, reliable view of RG-UMC on five datasets}\label{reliable-KL-UMC}
% \vspace{-0.3cm}
\begin{tabular}{l|c|ccc|ccc|ccc|ccc|ccc}
% \toprule[1pt]
\hline\hline
% \multirow{2}{*}{\textbf{dataset}}& \multirow{2}{*}{\textbf{strategy}}& \multicolumn{3}{c}{\emph{Original}} \vline &\multicolumn{3}{c}{\emph{Appoint}} \vline &\multicolumn{3}{c}{\emph{Random}} \vline
% &\multicolumn{3}{c}{\emph{Reliable(V-1)}}\vline
% &\multicolumn{3}{c}{\emph{Reliable(V)}}\\ \cline{3-17}
\multirow{2}{*}{\textbf{dataset}}& \textbf{strategy}& \multicolumn{3}{c}{\emph{Original}} \vline &\multicolumn{3}{c}{\emph{Appoint}} \vline &\multicolumn{3}{c}{\emph{Random}} \vline
&\multicolumn{3}{c}{\emph{Reliable(V-1)}}\vline
&\multicolumn{3}{c}{\emph{Reliable(V)}}\\ \cline{2-17}
&\textbf{view} &\emph{NMI} &\emph{ACC} &\emph{F1} &\emph{NMI} &\emph{ACC} &\emph{F1} &\emph{NMI} &\emph{ACC} &\emph{F1} &\emph{NMI} &\emph{ACC} &\emph{F1}&\emph{NMI} &\emph{ACC} &\emph{F1}\\\hline

% digit-5views
% \multirowcell{6}{\textbf{\emph{Digit}}\\\textbf{\emph{(5views)}}}
% &v1 &46.34	&47.25	&44.81	&90.60	&85.75	&83.40	&91.05	&83.75	&82.16	&87.75	&90.00	&90.16	&93.47	&88.50	&85.28\\
% &v2 &64.49	&56.25	&55.12	&86.15	&89.25	&89.50	&95.22	&97.50	&97.51	&90.08	&86.00	&83.62	&86.08	&81.75	&80.13\\
% &v3 &40.28	&50.25	&49.98	&85.72	&89.50	&89.82	&88.88	&89.25	&89.40	&96.08	&97.75	&97.77	&91.19	&94.25	&94.25\\
% &v4 &65.86	&67.25	&65.64	&89.79	&83.75	&81.82	&90.97	&92.75	&92.68	&93.60	&88.75	&85.50	&93.24	&92.75	&92.42\\
% &v5 &43.83	&41.00	&39.08	&87.96	&89.00	&89.27	&88.72	&84.00	&81.98	&89.65	&83.50	&82.15	&87.10	&87.50	&87.89\\
% & all-view &12.20	&18.15	&19.13	&89.35	&91.75	&91.95	&91.99	&94.80	&94.83	&93.75	&96.25	&96.25	&\textbf{94.05}	&\textbf{96.45}	&\textbf{96.45}\\\hline%\cline{2-17}

\multirowcell{7}{\textbf{\emph{Digit}}\\\textbf{\emph{(6views)}}}
&v1 &44.18	&47.27	&46.41	&85.49	&85.76	&86.18	&69.60	&72.12	&70.55	&69.77	&70.00	&68.49	&66.59 &66.36 &65.11\\
&v2 &64.49	&53.64	&51.68	&84.33	&83.94	&83.46	&89.73	&83.94	&82.36	&82.82	&87.58	&87.71	&84.04	&79.70	&77.58\\
&v3 &42.35	&48.79	&49.19	&62.77	&69.70	&70.75	&73.10	&82.42	&82.13	&87.38	&91.21	&90.56	&75.89	&84.55	&83.99\\
&v4 &61.98	&71.21	&71.61	&81.73	&80.61	&78.27	&87.69	&83.03	&81.52	&81.05 &76.97	&76.09	&85.30	&83.33	&80.52\\
&v5 &47.87	&47.88	&46.00	&67.16	&70.00	&68.87	&73.83	&76.06	&74.87	&72.59	&72.73	&70.37	&73.00	&72.73	&71.53\\
&v6 &67.37	&65.45	&61.75	&70.29	&70.00	&64.75	&69.48	&67.58	&63.19	&75.54	&75.76	&70.54	&70.85	&71.21	&66.24\\
&all-view&8.57	&15.30	&15.46	&72.24	&71.11	&66.18	&\textbf{77.93}	&{80.15}	&{82.57}	&76.26	&\underline{84.24}	&\underline{83.96}	&\underline{77.37}	&\textbf{84.85}	&\textbf{84.65}\\\hline%\cline{2-17}

\multirowcell{4}{\textbf{\emph{Scene15}}\\\textbf{\emph{(3views)}}}%\multirow{4}{*}{\textbf{\emph{Scene15(3views)}}}
&v1	&39.98	&36.27	&32.79	&46.23	&42.91	&41.52	&43.04	&37.41	&31.82	&46.93	&45.80	&42.26	&46.67	&43.12	&39.20\\
&v2	&38.20	&34.12	&31.92	&53.73	&48.82	&47.32	&50.65	&48.56	&47.34	&53.59	&49.50	&47.13	&53.98	&45.80	&42.73\\
&v3	&19.87	&22.57	&20.21	&38.41	&37.74	&32.85	 &32.48	&31.03	&28.50 	&38.47	&36.40	&31.34	&38.13	&35.59	&32.21\\
&all-view	&20.74	&19.21	&18.57	&\underline{48.15}	&\underline{53.28}	&\underline{48.85}	&42.81	&44.86	&42.12	&47.82	&51.20	&47.15	&\textbf{49.23}	&\textbf{53.82}	&\textbf{49.82}\\\hline%\cline{2-17}

% rand 
% &46.36	&40.43	&37.82
% &54.25	&45.40	&42.05
% &37.71	&34.25	&30.95
% &50.40	&54.40	&50.79
\multirowcell{7}{\textbf{\emph{Caltech101-20}}\\\textbf{\emph{(6views)}}}
&v1	&42.81	&33.68	&27.41	&57.38	&59.90	&33.31	&57.15	&44.47	&31.82	&57.68	&58.35	&37.40	&53.86	&53.73	&31.51\\
&v2	&45.15	&33.42	&27.46	&49.05	&43.44	&27.89	&57.75	&53.73	&35.41	&53.69	&46.53	&34.18	&54.96	&44.47	&32.31\\
&v3	&45.03	&36.76	&32.52	&48.86	&39.59	&27.69	&54.20	&44.47	&27.02	&56.04	&53.21	&30.07	&54.20	&42.67	&30.19\\
&v4	&59.86	&42.16	&31.22	&59.27	&46.53	&34.01	&66.00	&54.50	&36.82	&70.74	&49.10	&38.77	&67.99	&50.90	&36.77\\
&v5	&52.58	&38.82	&32.52	&56.68	&48.59	&33.04	&60.15	&46.79	&30.95	&64.07	&50.39	&32.04	&65.13	&47.56	&36.71\\
&v6	&59.56	&45.50	&31.07	&60.62	&53.21	&37.31	&65.12	&54.24	&41.25	&62.94	&46.53	&36.52	&62.62	&55.78	&34.33\\
&all-view	&19.72	&15.21	&11.08	&59.37	&\underline{64.27}	&32.59	&62.91	&61.74	&\underline{37.39}	&\underline{64.49}	&{63.62}	&{37.24}	&\textbf{65.81}	&\textbf{67.05}	&\textbf{38.92}\\\hline%\cline{2-17}

% \multirow{8}{*}{\textbf{flower17-7v}} &v1	&50.59	&38.50	&38.06	&51.74	&39.04	&38.58	&52.28	&39.57	&38.53	&53.99	&40.64	&39.47	&52.68	&40.64	&38.93\\
% &v2	&37.73	&26.20	&25.59	&44.26	&31.55	&31.24	&42.23	&32.09	&32.27	&42.38	&32.62	&33.01	&38.69	&28.88	&28.15\\
% &v3	&48.75	&37.97	&38.36	&55.11	&43.85	&41.26	&53.11	&42.78	&41.3	&52.70	&45.99	&44.94	&52.87	&41.71	&40.44\\
% &v4	&44.37	&32.09	&30.88	&43.31	&30.48	&30.56	&45.27	&33.16	&32.36	&45.82	&36.90	&36.62	&46.06	&35.29	&35.34\\
% &v5	&51.66	&42.25	&42.89	&53.10	&41.18	&40.11	&53.53	&40.64	&41.36	&51.75	&41.18	&41.74	&51.05	&37.97	&38.55\\
% &v6	&38.40	&28.34	&27.47	&43.47	&29.95	&28.98	&41.48	&27.81	&26.94	&40.23	&28.88	&27.69	&39.52	&26.74	&25.47\\
% &v7	&37.97	&26.74	&25.83	&40.95	&30.48	&29.25	&39.70	&29.41	&28.98	&40.33	&31.02	&29.71	&39.65	&28.88	&27.58\\
% & \xlk{all-view}	&11.86	&11.92	&11.93	&16.25	&15.13	&15.51	&16.28	&16.88	&16.12	&18.08	&16.81	&16.06	&51.25&43.77&42.11 \\\hline%\cline{2-17}
\multirowcell{8}{\textbf{\emph{Flower17}}\\\textbf{\emph{(7views)}}} &v1	&50.59	&38.50	&38.06	&59.73	&52.94	&52.25	&59.80	&50.27 &47.52	&66.16	&56.15	&54.55	&59.61	&50.80	&49.16\\
&v2	&37.73	&26.20	&25.59	&40.72	&32.62	&29.92	&40.18	&32.62	&30.00	&39.37	&31.55	&30.59	&41.34	&32.62	&30.52\\
&v3	&48.75	&37.97	&38.36	&55.63	&44.39	&42.41	&48.80	&41.18	&39.49	&53.80	&43.32	&37.20	&55.60	&43.32	&40.71\\
&v4	&44.37	&32.09	&30.88	&45.08	&37.43	&37.21	&44.11	&35.83	&35.17	&49.19	&37.43	&34.71	&50.08	&36.90	&33.40\\
&v5	&45.63	&34.76	&34.67	&42.49	&34.22	&28.95	&45.01	&39.57	&38.96	&50.00	&36.90	&34.06	&45.19	&36.90	&37.12\\
&v6	&39.81	&33.16	&31.34	&50.65	&37.97	&34.56	&55.48	&44.39	&42.05	&51.16	&38.50	&34.63	&45.47	&35.83	&32.54\\
&v7	&45.3	&34.76	&35.90	&45.97	&36.90	&35.30	&52.36	&42.78	&41.64	&48.78	&44.39	&45.20	&47.18	&37.43	&36.12\\
& {all-view}	&11.70&12.15&12.51	&{53.15}&45.07&40.81	&49.67&{48.20}&{46.27}	&\underline{56.46}&\underline{49.89}&\underline{46.84}	&\textbf{58.57}&\textbf{50.04}&\textbf{47.81}\\\hline%\cline{2-17}
%&\textbf{50.39}&\textbf{40.95}&\textbf{39.70} % &49.28&40.79&39.31	
\multirowcell{6}{\textbf{\emph{Reuters}}\\\textbf{\emph{(5views)}}} 
&v1	&14.60	&35.61	&27.73	&30.50	&46.84	&26.93	&35.67	&50.57	&36.30	&32.17	&51.16	&31.62	&27.28	&50.28	&22.58\\
&v2	&15.72	&35.02	&28.04	&37.60	&49.16	&23.69	&38.68	&63.32	&36.53	&33.16	&53.19	&35.95	&36.49	&62.47	&35.95\\
&v3	&14.38	&34.92	&27.06	&34.61	&44.79	&31.07	&36.00	&55.75	&34.64	&25.52	&49.80	&22.25	&24.54	&50.20	&22.10\\
&v4	&11.86	&33.08	&27.23	&34.95	&47.80	&31.41	&36.30	&54.63 &37.01	&27.15	&50.57	&22.49	&28.58	&50.73	&22.80\\
&v5	&14.50	&33.56	&27.32	&19.35	&41.05	&23.36	&19.68	&45.29	&31.47	&20.33	&53.45	&30.54	&20.25	&53.51	&30.62\\
&all-view	&3.95	&28.30	&13.78	&28.63	&46.89	&24.55	&30.70	&57.44	&\bf{34.73}	&\textbf{31.01}	&\underline{59.78}	&{34.52}	&\underline{30.93}	&\textbf{59.78}	&\underline{34.54}\\%\hline%\cline{2-17}
% reuters-5views &30.29	&58.75	&33.98
% \multirow{4}{*}{\textbf{ExtendYaleb-3v}}&v1	&27.83	&33.81	&34.22	&50.48	&54.29	&56.62	&40.18	&39.05	&38.60	&52.76	&48.10	&46.98	&49.15	&42.86	&42.28\\
% &v2	&28.13	&35.71	&35.94	&46.40	&37.62	&37.47	&40.98	&43.33	&42.40	&46.42	&42.86	&47.11	&49.80	&47.14	&49.03\\
% &v3	&9.96	&18.10	&17.86	&50.07	&43.81	&45.67	&39.43	&37.62	&36.27	&44.83	&37.62	&36.49	&50.63	&43.81	&41.42\\
% &all-view	&5.07	&15.71	&15.43	&48.73	&48.89	&46.75	&48.22	&45.40	&49.12	&56.05	&53.33	&51.87	&58.55	&50.95	&49.70\\%\cline{2-17}

\hline\hline
\end{tabular}
% \vspace{-0.1cm}
\end{table*}

\textbf{Influence of multiple reliable views in RGs-UMC model.}
In the RGs-UMC model, two variations of Eq. (\ref{eq16}) are introduced: uniform treatment and direct normalization. (i) Uniform treatment assigns equal weights of 1 to all selected reliable views. (ii) Direct normalization involves normalizing the silhouette coefficients of different views directly, which selects reliable views based on their normalized coefficients. These variations are formulated as:
\begin{align}
\setlength\abovedisplayskip{1.5pt}
\setlength\belowdisplayskip{1.5pt}
% \small
    wu_{vr} &= \left \{
        \begin{array}{ll}
        1,                    & \textit{{\rm if}}\quad  sils^{r} \geq sils^{v},\\
            % 1,                    & \textit{\rm {if}}\quad  \rm{sils^{r}} \geq  sils^{v},\\
            0,                    & \textit{\rm {otherwise.}}\\
        \end{array}
        \right.\label{eq14}
    \\
    wn_{vr} &= \left \{
    \begin{array}{ll}
        sils^r/\sum_{i=1}^{|\Omega_R|} sils^i,                    & \textit{{\rm if}}\quad  sils^{r} \geq sils^{v},\\
        % sils^r/\sum_{i=1}^{|\Omega_R|} sils^i,                    & \textit{\rm {if}}\quad  \rm{sils^{r}} \geq  sils^{v},\\
        0,                    & \textit{{\rm otherwise.}}\\
    \end{array}
    \right.\label{eq15}
\end{align}
where the symbol meaning in Eq. (\ref{eq14}) and Eq. (\ref{eq15}) is consistent with that in Eq. (\ref{eq16}). The two approaches for assigning weights to multiple reliable views in RGs-UMC are referred to as URGs-UMC (Uniform RGs-UMC) and NRGs-UMC (Normalization RGs-UMC), respectively. 
The results are presented in TABLE \ref{table5_multi_view_morethantwo_2} compared with RGs-UMC. 

From TABLE \ref{table5_multi_view_morethantwo_2}, we conclude that: 1) URGs-UMC treats each view equally, which diminishes the guiding effect of views with better cluster structures. 2) NRGs-UMC diminishes the guiding effect of views with smaller silhouette coefficients, making them almost ineffective in the clustering process. 3) It is evident that the performance of RGs-UMC is relatively stable compared to URGs-UMC and NRGs-UMC methods. In RGs-UMC, it uses a sigmoid function to balance the weights, ensuring that each view contributes meaningfully to the clustering process.

\begin{table*}%[t]
\centering
%\vspace{-1mm}
\renewcommand{\arraystretch}{1.05}
\setlength{\tabcolsep}{5.5pt}
\caption{Comparison of the different weight strategies (Uniform Treatment and Direct Normalization). }\label{table5_multi_view_morethantwo_2}
% \vspace{-0.3cm}
\begin{tabular}{l|ccc|ccc|ccc|ccc|ccc}
%\toprule[1pt]
\hline\hline
\multirow{2}{*}{\textbf{Methods}}& \multicolumn{3}{c}{\emph{Digit (6views)}} \vline &\multicolumn{3}{c}{\emph{Scene15 (3views)}} \vline &\multicolumn{3}{c}{\emph{Caltech101-20 (6views)}} \vline
&\multicolumn{3}{c}{\emph{Flower17 (7views)}}\vline&\multicolumn{3}{c}{\emph{Reuters (5views)}}\\ \cline{2-16}
&\emph{NMI} &\emph{ACC} &\emph{F1} &\emph{NMI} &\emph{ACC} &\emph{F1} &\emph{NMI} &\emph{ACC} &\emph{F1} &\emph{NMI} &\emph{ACC} &\emph{F1}
&\emph{NMI} &\emph{ACC} &\emph{F1}\\\hline
\textbf{URGs-UMC}&\underline{84.01}	&\underline{90.96}	&\underline{90.87}    &42.01	&48.02	&46.04    &\underline{65.42}	&\underline{73.35}	&\textbf{41.67}    &\underline{55.02}	&\underline{49.50}	&\underline{49.35}    &22.03	&46.16	&31.52\\
% digit-5views 
%\textbf{URGs-UMC}  &97.86	&98.80	&98.80
\textbf{NRGs-UMC}&80.86	&88.38	&88.18   &\underline{45.10}	&\underline{51.47}	&\underline{46.66}    &61.09	&62.25	&33.74    &53.41	&48.05	&45.61     &\underline{27.30}	&\underline{54.71}	&\underline{32.12}\\
% digit-5views 
%\textbf{NRGs-UMC} &97.79	&98.95	&98.95 
\textbf{RGs-UMC} &\textbf{86.19}&\textbf{92.83}&\textbf{92.77}       &\textbf{47.40}&\textbf{53.50}&\textbf{48.11}    &\textbf{70.88}&\textbf{75.28}&\underline{41.50}    &\bf{60.64}&\bf{52.02}&\bf{51.68} &\textbf{32.26}&\textbf{57.24}&\textbf{33.35}\\
% digit-5views
%\textbf{RGs-UMC} &\textbf{97.95}&\textbf{99.05}&\textbf{99.05}
\hline\hline
\end{tabular}
\vspace{-0.1cm}
\end{table*}

% \textbf{The performance of KL-divergence module as plug-and-play unit.}
% \begin{table}
% \vspace{-0.1cm}
% \centering
% %\small
% \renewcommand{\arraystretch}{1.0}
% \setlength{\tabcolsep}{7pt}
% \caption{KL as plug-and-play unit on \emph{Caltech101-20} with two views}\label{}
% \vspace{-0.3cm}
% \begin{tabular}{c|ccc}
% %\toprule[1pt]
% \hline\hline
% Caltech101 & \emph{NMI} & \emph{ACC} & \emph{F1}\\
% \hline
% completer& 33.77& 26.31& 21.38\\%\hline
% completer+KL& 36.85 &29.25 &24.36\\%\hline
% \hline\hline
% \end{tabular}
% \vspace{-0.4cm}
% \end{table}

\begin{figure}%[!t]
\centering
\includegraphics[width=1.0\columnwidth]{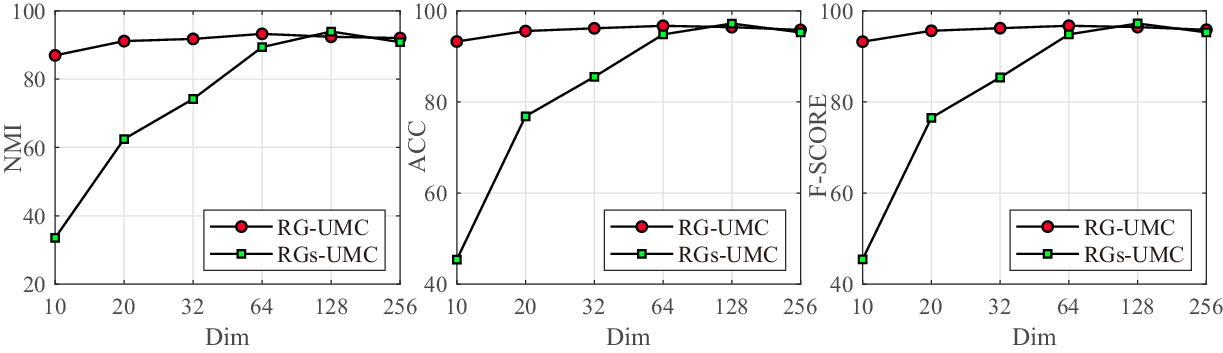}% OK
% \vspace{-0.1cm}
\caption{Influence of the latent dimension.}
% \caption{The influence of reliable views selection and latent dimension.}
\label{KL_WKL_UMC_latent}
% \vspace{-0.1cm}
\end{figure}

\textbf{Influence of latent dimension.}
% The last layer of autoencoder is softmax layer as the table \ref{table2_networkarchitecture} shows. 
The last layer of the autoencoder is the softmax layer as the network architecture shows \cite{scl-UMC}. We treat each element of the representation as an over-cluster class probability like \cite{lin2021completer}. Take the two-view \emph{Digit} dataset for example, we change the dimension of the latent representation in the range of $\{K, 2K, 32, 64, 128, 256\}$, where K(10) is the class number of the \emph{Digit} dataset. Fig. \ref{KL_WKL_UMC_latent} displays the performance changes of the RG-UMC and RGs-UMC with varying dimensions. As Fig. \ref{KL_WKL_UMC_latent} shows, both models achieve good performance when the dimension is greater than 64. Therefore we set the dimension of latent representation with 128 in RG-UMC and RGs-UMC models, which is consistent with the setting of the completer \cite{lin2021completer}.

\begin{figure*}[!t]
\centering
\includegraphics[width=1.8
\columnwidth]{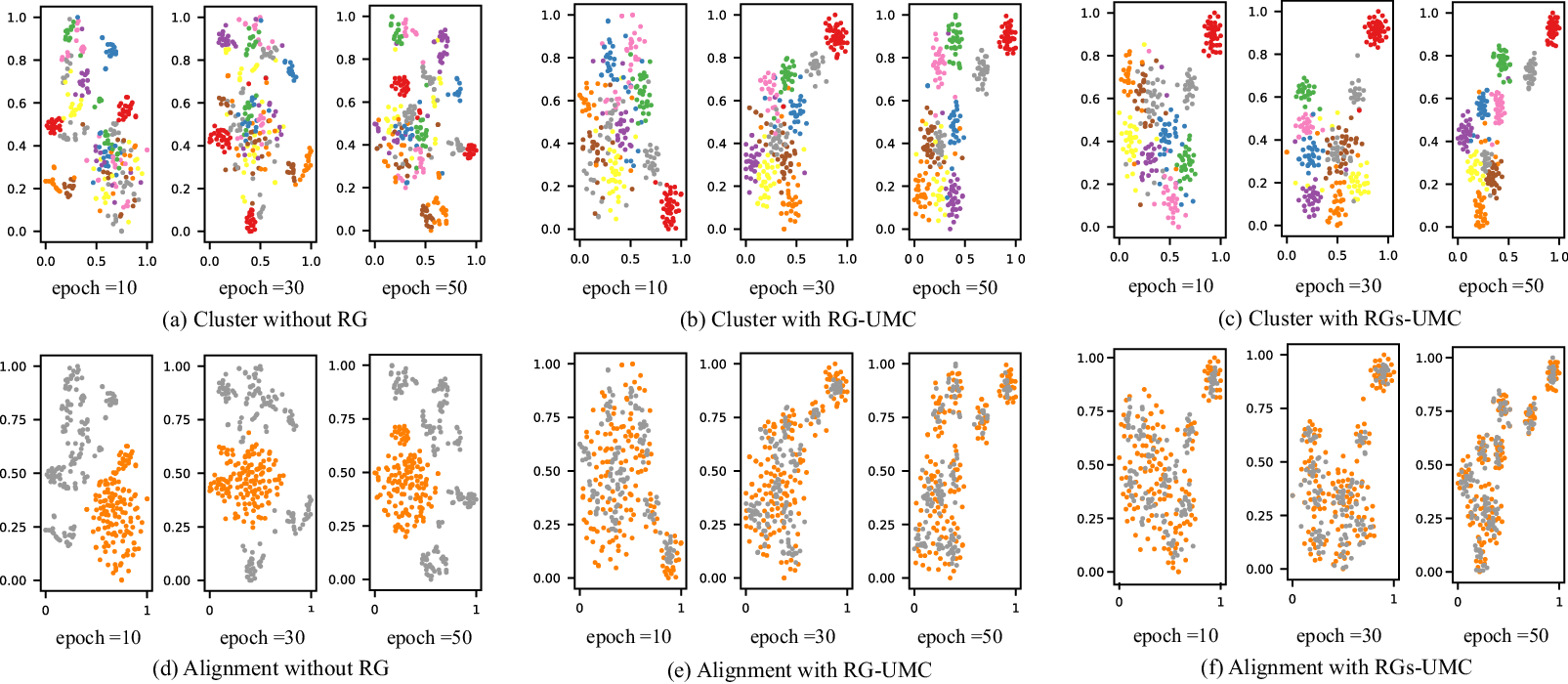}
\vspace{-0.1cm}
\caption{Evolution of latent representations for cluster formation and views alignment during training, every twenty epochs, on the two-view \emph{Digit} dataset. }\label{KL-WKL-UMC-singlecluster}
\end{figure*}

\textbf{Visualization.} 
% \xlk{To illustrate the effectiveness of learning a consistent cluster structure, we utilize t-SNE visualization to observe the evolution of the latent subspace every twenty epochs during the training process conducted on the \emph{Digit} dataset with two views.}
{To illustrate the effectiveness of learning a consistent cluster structure, the t-SNE visualization is utilized to observe the latent subspace evolution every twenty epochs during the training, conducted on the two-view \emph{Digit} dataset.}
{For clear illustration, we randomly sampled 300 instances of $\boldsymbol{Z}$ and illustrated the change evolution of them in Fig. \ref{KL-WKL-UMC-singlecluster}.} In Fig. \ref{KL-WKL-UMC-singlecluster} (a-c), color represents cluster assignments predicted by the $K$-means algorithm. Besides, in Fig. \ref{KL-WKL-UMC-singlecluster} (d-e), color denotes different views. Fig. \ref{KL-WKL-UMC-singlecluster} (a) and (d) depict the model in this paper without the alignment module. From these visualizations, {we conclude that:}

{1) The effectiveness of reliable view guidance is demonstrated by comparing Fig. \ref{KL-WKL-UMC-singlecluster} (a) and (d) with Fig. \ref{KL-WKL-UMC-singlecluster} (b-c) and (e-f).}
{2) Observing Fig. \ref{KL-WKL-UMC-singlecluster} (b-c), we note the initial mixing of features and the sparse appearance of clusters.} However, with training, the cluster assignments become more rational, and features tend to gather and scatter more distinctly. This observation confirms the effect of the RG-UMC and RGs-UMC models in learning clear cluster structures. 
3) Fig. \ref{KL-WKL-UMC-singlecluster} (e-f) shows that our methods successfully align the samples from different views during the training process. That is KL-divergence we used effectively aligns the distribution between reliable views and other views, and improves the clustering performance. 

\begin{figure}
%\vspace{-0.2cm}
\centering
\includegraphics[width=1.0\columnwidth]{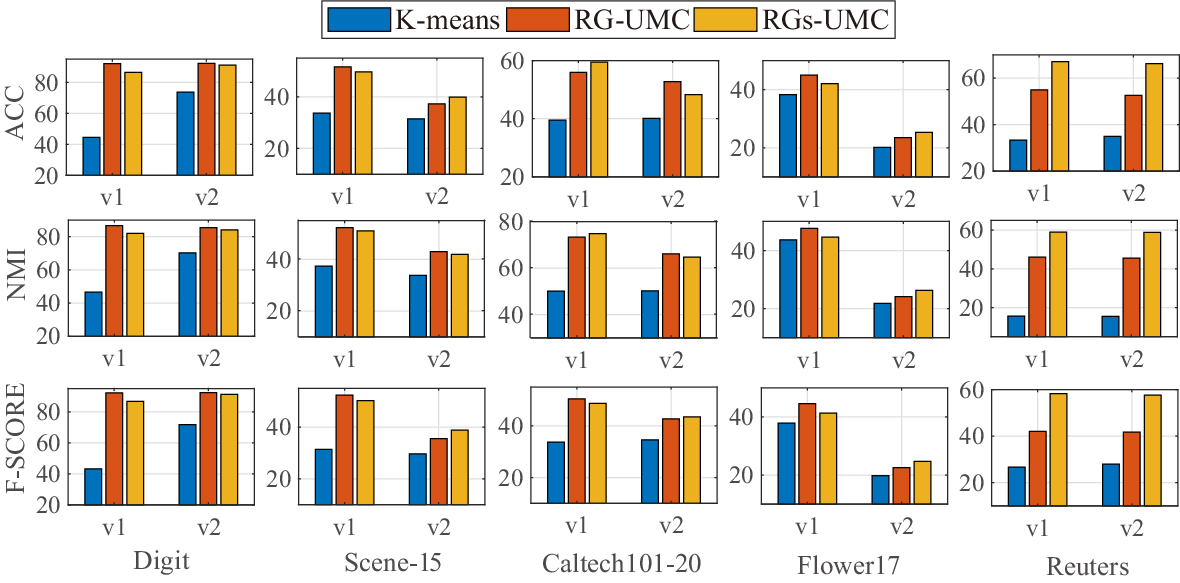}% change name
% \vspace{-0.1cm}
\caption{Performance comparison of observed samples in each view among $K$-means, RG-UMC and RGs-UMC.}\label{singleview_before_after}
\vspace{-0.2cm}
\end{figure}

\textbf{The comparison of clustering performance within individual views.} % 
{In multi-view clustering, clustering can be performed independently within each view. }Our method utilizes reliable views to guide the clustering of other views, enabling joint clustering across multiple views. Therefore, it is necessary to explore the changes in single-view clustering performance. Then we evaluated the performance of the RG-UMC and RGs-UMC models on each view across five two-view datasets with $K$-means. 
{As depicted in Fig. \ref{singleview_before_after}, our methods significantly improve the clustering performance within each individual view.} The improvement can be attributed to the reliable information conveyed by other views through the reliable alignment module via KL divergence. %subspace representation.

\section{Conclusion and Future Work}\label{conclusions}
UMC is a challenging task as it lacks supervision information and matching samples between views. To establish relationships between different views, we utilize reliable view guidance to explore the consistent and complementary information of cluster structure between views. Then, two novel models, RG-UMC and RGs-UMC, both of which dynamically leverage the reliable view to guide the learning process, are proposed.
{Our models effectively address the challenges posed by uncertain cluster structures and pairing relationships between views. Through comprehensive experiments, we validate the effectiveness of our methods, showing remarkable performance in UMC.} 

{In our work, we leverage a consistent cluster structure to establish connections. In the future, we plan to explore deeper relationships between views through hierarchical pairing. Moreover, the reliability and security of UMC are important. Therefore, it is crucial to explore algorithms within a stable and secure framework. Additionally, we need to thoroughly investigate algorithms tailored to the specific applicability of UMC.} 
% multi-layer or multi-scale learning

\bibliographystyle{IEEEtran}
\bibliography{IEEEabrv,egbib1}

\end{document}